\documentclass[11pt]{article}
\usepackage[margin=1.25in]{geometry}
\usepackage{amsmath,amsfonts,amsthm,amssymb}
\usepackage{algpseudocode}
\usepackage{algorithm}
\usepackage{footnote}
\usepackage{color,xcolor,colortbl}
\definecolor{LightCyan}{rgb}{0.88,1,1}

\usepackage{hyperref}

\usepackage{authblk}
\usepackage{natbib}

\usepackage[english]{babel}

\usepackage{graphicx,subfigure}
\usepackage[ulem=normalem]{changes}

\usepackage[title]{appendix}

\usepackage{times}

\usepackage{url}

\usepackage{todonotes}

\numberwithin{equation}{section}

\title{A Theoretical Analysis of Noisy Sparse Subspace Clustering on Dimensionality-Reduced Data
\footnote{A shorter version of this paper titled ``A Deterministic Analysis of Noisy Sparse Subspace Clustering on Dimensionality-Reduced Data'' with partial results appeared at \emph{Proceedings of the 32nd International Conference on Machine Learning (ICML)} held at Lille, France in 2015.}
}
\author{Yining Wang}
\author{Yu-Xiang Wang}
\author{Aarti Singh}
\affil{Machine Learning Department, Carnegie Mellon University, USA\\
{\{yiningwa,yuxiangw,aarti\}@cs.cmu.edu}}

\newcommand{\vct}{\boldsymbol }
\newcommand{\mat}{\mathbf}

\newcommand{\nml}{\mathcal{N}}

\newcommand{\argmin}{\mathrm{argmin}}

\newcommand{\conv}{\mathrm{conv}}
\newcommand{\E}{\mathbb{E}}

\newcommand{\SPAN}{\mathrm{span}}
\newcommand{\affn}{\mathrm{aff}_{\mathrm{N}}}

\def\R{\mathbb{R}}

\def\cA{\mathcal{A}}

\def\cN{\mathcal{N}}
\def\cO{\mathcal{O}}

\def\cS{\mathcal{S}}

\def\cX{\mathcal{X}}

\newtheorem{thm}{Theorem}[section]
\newtheorem{lem}{Lemma}[section]
\newtheorem{cor}{Corollary}[section]
\newtheorem{prop}{Proposition}[section]

\newtheorem{defn}{Definition}[section]

\allowdisplaybreaks[4]

\begin{document}

\maketitle
\begin{abstract}
	%\todo{YX: This is much different from what we had in the conference version. It's fine to include a bit more details, but I would request some rewording here that makes it more structured.}
Subspace clustering is the problem of partitioning unlabeled data points into a number of \emph{clusters}
so that data points within one cluster lie approximately on a \emph{low-dimensional linear subspace}.
In many practical scenarios, the dimensionality of data points to be clustered are compressed due to constraints of measurement, computation or privacy.
In this paper, we study the theoretical properties of a popular subspace clustering algorithm named \emph{sparse subspace clustering (SSC)}
and establish formal success conditions of SSC on dimensionality-reduced data.
Our analysis applies to the most general fully deterministic model where both underlying subspaces and data points within each subspace are deterministically positioned,
and also a wide range of dimensionality reduction techniques (e.g., Gaussian random projection, uniform subsampling, sketching) that fall into
 %\todo{YW: Should we cite Woodruff et. al. for ``subspace embedding''? }
 a \emph{subspace embedding} framework \citep{lsr-ose,tensor-sketch-ose}.
Finally, we apply our analysis to a differentially private SSC algorithm and established both privacy and utility guarantees of the proposed method.
\end{abstract}

\section{Introduction}
%\todo{YX: This is arguably the most important part of the paper. I tried my best to edit it, but it still seems a little unsatisfactory. I will revisit this after Aarti's pass.}

Subspace clustering is an unsupervised learning paradigm aiming at grouping unlabeled data points into disjoint \emph{clusters} so that data points within each cluster lie near a \emph{low-dimensional linear subspace}. It has found many successful applications in computer vision and machine learning,
as many high dimensional data can be approximated by a union of low-dimensional subspaces.
Examples include motion trajectories \citep{sc-motion-trajectory}, face images \citep{sc-face-images},
network hop counts \citep{high-rank-matrix-completion}, movie ratings \cite{sc-movie-rating} and social graphs \citep{sc-social-graph}.

%A large body of research has been devoted to subspace clustering in the last two decades.

The success in applications is made possible by two decades of algorithmic research on this problem. Popular approaches include Expectation-Maximization-style methods such as K-plane \citep{kplane} and Q-flat \citep{qflat},
algebraic methods such as generalized principal component analysis \citep{gpca},
matrix factorization methods \citep{sc-motion-trajectory},
bottom-up local affinity-based methods such as those proposed by \citet{lsa,alc},
and convex optimization based approaches including Low Rank Representation (LRR, \citealp{lrr})
and Sparse Subspace Clustering (SSC, \citealp{ssc}).

In this paper we consider the SSC algorithm, which has drawn much interest from the literature. It is known that SSC enjoys superb performance in practice \citep{ssc}
and has theoretical guarantees under fairly general conditions \citep{sc-geometric,noisy-ssc,robust-ssc}.
Let $\mat X\in\mathbb R^{d\times N}$ denote the data matrix, where $d$ is the ambient dimension and $N$ is the number of data points.
For noiseless data (i.e., data points lie exactly on low-rank subspaces),
the exact SSC algorithm solves %the optimization problem in Eq. (\ref{eq_ssc}) 
\begin{equation}
\min_{\vct c_i\in\mathbb R^N}\|\vct c_i\|_1,\quad
\text{s.t. }  \vct x_i=\mat X\vct c_i, \vct c_{ii} = 0.
\label{eq_ssc}
\end{equation}
for each data point $\vct x_i$ to obtain self regression solutions $\vct c_i\in\mathbb R^N$.
For noisy data, the following Lasso version of SSC is often used in practice:
\begin{equation}
\min_{\vct c_i\in\mathbb R^N}\|\vct x_i-\mat X\vct c_i\|_2^2 + 2\lambda\|\vct c_i\|_1, \quad\text{s.t. }\vct c_{ii} = 0.
\label{eq_lasso_ssc}
\end{equation}
The intuition of SSC is to learn a sparse ``self-representation'' matrix $\mat C\in\mathbb R^{N\times N}$ over all data points by imposing an $\ell_1$ penalty on the representation coefficients. This can also be thought of as a convex optimization based graph embedding that captures a specific type of relationship among data points. Each non-zero entry of $\mat C$ is an edge connecting two data points. 
It has been shown in \cite{sc-geometric,noisy-ssc,robust-ssc} that under mild conditions the learned representation/similarity matrix $\mat C$ contains \emph{no false connections},
in that every such edge connects only data points that belong to the same cluster.
% non-zero element in $\mat C$ corresponds to data points that belong to the same cluster.
Finally, spectral clustering \citep{spectral-clustering} is performed on the learned similarity matrix $|\mat C|+|\mat C|^T$ to cluster the $N$ data points into disjoint clusters.

% explain why dimensionality-reduced data is prevalent. Say some examples (e.g., random Gaussian (compressed measurement), missing data, privacy, etc.)

Although success conditions for both exact SSC and Lasso SSC have been extensively analyzed in previous literature,
in practice it is inefficient or even infeasible to operate on data with high dimension.
Some types of dimensionality reduction is usually required \citep{vidal2010tutorial}.
In this paper, we propose a theoretical framework that analyzes SSC under many popular dimensionality reduction settings, including
\begin{itemize}
\item \textbf{Compressed measurement}: With compressed measurement dimensionality-reduced data are obtained by multiplying the original data
typically with a random Gaussian matrix.
%It is widely used in the field of compressed sensing and sparse recovery.
We show that SSC provably succeeds when the projected dimension is on the order of a low-degree polynomial of the maximum intrinsic rank of each subspace.

\item \textbf{Efficient computation}: By using fast Johnson-Lindenstrauss transform (FJLT) \citep{fjlt}
or sketching \citep{count-sketch,cs-ose}
one can reduce the data dimension for computational efficiency
while still preserving important structures in the underlying data.
We prove similar results for both FJLT and sketching.

\item \textbf{Handling missing data}: In many applications, the data matrix may be incomplete due to measurement and sensing limits.
It is shown in this paper that, when data meet some incoherent criteria, uniform feature sampling suffices for SSC.

\item \textbf{Data privacy}: Privacy is an important concern in modern machine learning applications.
It was shown that Gaussian random projection with added Gaussian noise preserves both information-theoretic \citep{zhou2009compressed} and differential privacy \citep{data-privacy-jl}. We provide a utility analysis which shows that SSC can achieve \emph{exact} subspace detection despite stringent privacy constraints.
%Consequently, our analysis framework provides a simple way of performing subspace clustering while preserving privacy.
\end{itemize}

% contribution: a fully deterministic analysis. why it's difficult? (standard SSC does not have a strongly convex dual problem)

The main contribution of this paper is a unified framework for analyzing sparse subspace clustering on dimensionality-reduced data.
In particular, we prove that a \emph{subspace embedding} property \citep{cs-ose,tensor-sketch-ose}
is sufficient to guarantee successful execution of the SSC algorithm after compression.
Furthermore, the lowest dimension we can compress the data into only scales as a low-degree polynomial of the \emph{intrinsic} dimension $r$
and does not depend (up to poly-logarithmic factors) on either the \emph{ambient} dimension $d$ or the total number of data points $N$.
This is a much desired property, because in practical subspace clustering applications both $d$ and $N$ are huge, while the intrinsic dimensionality $r$ typically stays at nearly a constant \citep{vidal2010tutorial,noisy-ssc}.
We also show by simple derivations and citing existing results that all of the above-mentioned data compression schemes (Gaussian projection, uniform subsampling, FJLT, sketching)
are subspace embeddings, and hence fall into the perturbation analysis framework we formulated.
Finally, as an application of our analysis, we propose a differentially private SSC algorithm by random projection followed by Gaussian perturbation 
and prove both privacy and utility guarantees of the proposed algorithm.

%A key observation is that all projections for the aforementioned settings
%are \emph{subspace embeddings}, which means they uniformly preserve the two norm of any vector belonging to a low-rank subspace.
%Our analysis applies to the \emph{fully deterministic} setting under which both subspaces and data points within each subspace are placed deterministically.
%It can also handle data corrupted by deterministic or stochastic noise.
%This generalizes previous work \cite{sc-dr-semirandom} which only applies to semi-random models with noiseless data
%\footnote{
%In semi/fully random models the underlying subspaces and/or data points are distributed uniformly at random.
%Detailed definitions can be found in \cite{sc-geometric}.}.
%The fully deterministic setting poses more challenges because the perturbation of dual directions introduced in \cite{sc-geometric} cannot be easily bounded
%if exact SSC is used.
%As a result, even for noiseless data, we employ a Lasso SSC formulation to obtain strong convexity in the dual problem.

\subsection{Problem setup and notations}\label{subsec:setup}

\paragraph{Notations}
For a vector $\vct x$, $\|\vct x\|_p=(\sum_i{|\vct x_i|^p})^{1/p}$ denotes the vector $p$-norm of $\vct x$.
For a matrix $\mat A$, $\|\mat A\|_p=\sup_{\vct x\neq 0}\|\mat A\vct x\|_p/\|\vct x\|_p$ denotes the operator $p$-norm of $\mat A$.
In particular, $\|\mat A\|_2=\sigma_{\max}(\mat A)$ is the spectral norm of $\mat A$.
The subscript $p$ is sometimes dropped for $p=2$; that is, $\|\vct x\|=\|\vct x\|_2$ and $\|\mat A\|=\|\mat A\|_2$.

The uncorrupted data matrix is denoted as $\mat Y\in\mathbb R^{d\times N}$,
where $d$ is the ambient dimension and $N$ is the total number of data points.
$\mat Y$ is normalized so that each column has unit two norm, which does not alter the union of subspaces on which $\mat Y$ lies.
Each column in $\mat Y$ belongs to a union of $k$ subspaces $\mathcal U^{(1)}\cup\cdots\cup\mathcal U^{(k)}$.
For each subspace $\mathcal U^{(\ell)}$ we write $\mat Y^{(\ell)}=(\vct y_1^{(\ell)},\cdots,\vct y_{N_\ell}^{(\ell)})$ for all columns belonging to $\mathcal U^{(\ell)}$,
where $N_\ell$ is the number of data points in $\mathcal U^{(\ell)}$ and $\sum_{\ell=1}^k{N_\ell} = N$.
We assume the rank of the $\ell$th subspace $\mathcal U^{(\ell)}$ is $r_\ell$ and
define $r=\max_\ell{r_\ell}$.
In addition, we use $\mat U^{(\ell)}\in\mathbb R^{d\times r_\ell}$ to represent an orthonormal basis of $\mathcal U^{(\ell)}$.
The observed matrix is denoted by $\mat X\in\mathbb R^{d\times N}$.
Under the noiseless setting we have $\mat X=\mat Y$;
for the noisy setting we have $\mat X=\mat Y+\mat Z$
where $\mat Z\in\mathbb R^{d\times N}$ is a noise matrix which can be either deterministic or stochastic.

We use ``$-i$'' subscript to denote all except the $i$th column in a data matrix.
For example, $\mat Y_{-i} = (\vct y_1,\cdots,\vct y_{i-1},\vct y_{i+1},\cdots,\vct y_{N})$ and
$\mat Y_{-i}^{(\ell)} = (\vct y_1^{(\ell)},\cdots,\vct y_{i-1}^{(\ell)}, \vct y_{i+1}^{(\ell)}, \cdots,\vct y_{N_\ell}^{(\ell)})$.
For any matrix $\mat A$, let $\mathcal Q(\mat A) = \conv(\pm\vct a_1,\cdots,\pm\vct a_N)$
denote the symmetric convex hull spanned by all columns in $\mat A$.
For any subspace $\mathcal U$ and vector $\vct v$, denote $\mathcal P_{\mathcal U}\vct v = \argmin_{\vct u\in\mathcal U}{\|\vct u-\vct v\|_2}$
as the projection of $\vct v$ onto $\mathcal U$.

\paragraph{Models}
%Explain semi-random and fully random models here. Also explain the difference of stochastic and adversarial noise.
We consider three models for the uncorrupted data $\mat Y$ of increasing strictness of assumptions.
Such hierarchy of models were first introduced in \cite{sc-geometric} and have served as reference models in existing analysis of SSC methods \citep{robust-ssc,tsc,noisy-ssc,greedy-ssc}:
\begin{itemize}
\item \textbf{Fully deterministic model}: in the fully deterministic model both the underlying low-rank subspaces $\mathcal U^{(1)},\cdots,\mathcal U^{(k)}$ 
%\todo{I changed most $L$ into $k$ to be consistent with the notations. I might have missed some though..}
and the data points $\vct y_1^{(\ell)}, \cdots,\vct y_{N_\ell}^{(\ell)}$ in each subspace are deterministically placed.
This is the most general model for subspace clustering (except the model-free agnostic settings considered in the projective clustering literature \citep{tiny-data})
as no stochastic or i.i.d.~type assumptions are imposed on either the subspaces or the data points.

\item \textbf{Semi-random model}: In the semi-random model the underlying subspaces $\mathcal U^{(1)},\cdots,\mathcal U^{(k)}$ are again deterministically placed;
however, (uncorrupted) data points within each subspace are assumed to be sampled i.i.d.~\emph{uniformly} at random from the unit sphere of the corresponding low-dimensional subspace.
One advantage of semi-random modeling is its interpretability: success conditions of SSC could be fully characterized using affinities between subspaces
and number of data points per subspace \citep{sc-geometric,robust-ssc,noisy-ssc}.

\item \textbf{Fully-random model}: In the fully random model both the underlying subspaces and data points within each subspace are sampled uniformly at random.
\end{itemize}

Apart from data models, we also consider two models when data $\mat Y$ are corrupted by noise $\mat Z=(\vct z_1,\cdots,\vct z_N)\neq\mat 0$.
In the \emph{deterministic noise} model, each noise vector $\vct z_i$ is adversarially placed, except that the largest maximum magnitude $\max_{1\leq i\leq n}\|\vct z_i\|_2^2$
is upper bounded by a noise parameter $\xi$.
In the \emph{stochastic noise} model, $\vct z_i$ is assumed to be i.i.d.~sampled from a zero-mean multivariate distribution $\nml_d(\vct 0,\sigma^2/d\cdot \mat I_{d\times d})$
for some noise parameter $\sigma$.
Note that here we divide the variance $\sigma^2$ by $d$ to keep the magnitude of noise constant and not increasing with number of dimensions.
In our analysis, the stochastic noise model allows for larger magnitude of noise compared to the deterministic (adversarial) noise setting
as it places stronger assumptions on the properties of the noise. 

\paragraph{Methods}
The first step is to perform dimensionality reduction on the observation matrix $\mat X$.
More specifically, for a target projection dimension $p< d$,
the projected observation matrix $\widetilde{\mat X}'\in\mathbb R^{p\times N}$
is obtained by first computing $\widetilde{\mat X} = \mat\Psi\mat X$ for some random projection matrix $\mat\Psi\in\mathbb R^{p\times d}$
and then normalizing it so that each column in $\widetilde{\mat X}'$ has unit two norm.
Afterwards, Lasso self-regression as formulated in Eq.~(\ref{eq_lasso_ssc}) is performed for each column in $\widetilde{\mat X}'$
to obtain the similarity matrix $\mat C=\{\vct c_i\}_{i=1}^N\in\mathbb R^{N\times N}$.
Spectral clustering is then be applied to $\mat C$ to obtain an explicit clustering of $\mat X$.
In this paper we use the normalized-cut algorithm \cite{normalized-cut} for spectral clustering.

\paragraph{Evaluation measures}
To evaluate the quality of obtained similarity matrix $\mat C$, we consider the \emph{Lasso subspace detection property} defined in \cite{noisy-ssc}.
More specifically, $\mat C$ satisfies Subspace Detection Property (SDP) if for each $i\in\{1,\cdots,N\}$ the following holds:
1) $\vct c_i$ is a non-trivial solution. That is, $\vct c_i$ is not the zero vector;
2) if $c_{ij}\neq 0$ then data points $\vct x_i$ and $\vct x_j$ belong to the same subspace cluster.
The second condition alone is referred to as ``Self-Expressiveness Property'' (SEP) in \cite{ssc}.
Note that we do \emph{not} require $c_{ij}\neq 0$ for \emph{every} pair of $\vct x_i,\vct x_j$ belonging to the same cluster.
We also remark that in general SEP is not necessary for spectral clustering to succeed, cf. \cite{noisy-ssc} 
\footnote{It is almost sufficient for perfect clustering both in practice~\citep{ssc} and in theory~\citep{consistent-ssc}.}.
%\footnote{It is also not sufficient for perfect clustering due to possible graph connectivity issues \citep{nasihatkon2011graph}, but in practice it does often imply perfect clustering.}.

\subsection{Related Work}

\cite{sc-dr-semirandom} analyzed SSC and threshold-based subspace clustering (TSC, \citealp{tsc}) on dimension-reduced noiseless data.
\cite{sc-dr-semirandom-journal}, which was arxived concurrently with the earlier version in ICML of our paper,
 further generalized \cite{sc-dr-semirandom}, with analysis of TSC and SSC-OMP on noisy data.
One important limitation of both \cite{sc-dr-semirandom,sc-dr-semirandom-journal} is that the data points in each subspace is assumed to be 
drawn \emph{uniformly} at random, corresponding to the semi-random model specified in Sec.~\ref{subsec:setup}.
Though amenable for theoretical analysis, the semi-random model deviates significantly from subspace clustering practice because data points,
even on the same subspace, are not distributed uniformly at random.
This is particularly true when data lies on affine subspaces and the ``homogeneous embedding'' trick \footnote{Appending $1$ to every data point.} is used, which is almost always the case in practice \citep{ssc}.
The added dimension in homogeneous embedding is constant valued and breaks any semi-random assumptions.
In this paper, we complement the results of \citet{sc-dr-semirandom-journal} by analyzing noisy SSC under the \emph{fully deterministic} model,
where no stochastic assumptions imposed upon the data points.
Our analysis naturally leads to interpretable noise conditions under the semi-random model.

Our proof technique also significantly differs from the one in \cite{sc-dr-semirandom-journal} which focused primarily on perturbation of \emph{subspace affinities} \citep{sc-geometric}.
When data points are not uniformly distributed, subspace affinity as defined by \cite{sc-geometric,sc-dr-semirandom-journal} no longer serves as a good characterization of the difficulty of the subspace clustering problem.
Instead, we propose novel perturbation analysis of the dual solution of the noisy SSC formulation, which is applicable under the fully deterministic setting.
Finally, we remark that an earlier conference version of this paper \citep{sc-random-icml} which summarized most parts of the deterministic analysis in this paper (including analysis for noisy data)
was published before \cite{sc-dr-semirandom-journal}.
Another difference, as noted in \cite{sc-dr-semirandom-journal},
is that in our analysis the projected noise is added after normalization of the projected signal.
We consider this to be a minor difference because the length of the projected signal is close to the length of the original signal, thanks to Proposition \ref{prop_wellbehave_unisample}.
In particular, for semi-random or fully-random models the length of each data point is very close to one with high probability,
and hence the noisy model we analyzed behaves similarly to the one considered in \cite{sc-dr-semirandom-journal}.
%Our noise model is slightly different in that projected noise is added after normalization of the projected signal, as remarked in \cite{sc-dr-semirandom-journal}.
%However, we consider this to be a minor difference, especially under the semi-random setting because the norm of the projected signal is close to one anyway, thanks to Proposition \ref{prop_wellbehave_unisample}.

% NIPS paper: independent subspaces
Arpit et al.~proposed a novel dimensionality reduction algorithm to preserve independent subspace structures~\cite{indep-subspace-jl}.
They showed that by using $p=2k$, where $k$ is the number of subspaces, one can preserve the independence structure among subspaces.
However, their analysis only applies to noiseless and independent subspaces,
while our analysis applies even when the least principal angle between two subspaces diminishes
and can tolerate a fair amount of noise.
Furthermore, in our analysis the target dimension $p$ required depends on the maximum intrinsic subspace dimension $r$ instead of $k$.
Usually $r$ is quite small in practice \citep{ssc,sc-face-images}.

% high rank matrix completion

Another relevant line of research is \emph{high-rank matrix completion}.
In \cite{high-rank-matrix-completion} the authors proposed a neighborhood selection based algorithm to solve multiple matrix completion problems.
%Although their method does recover points lying on the same subspace,
%the completion problem is quite different from subspace clustering.
%For example, passive matrix completion algoithms require both row and column spaces of a matrix to be incoherent \citep{nc-bernstein-2,adaptive-matrix-completion-arxiv},
%while for subspace clustering we only assume incoherence on the column space.
%By considering the simpler clustering problem, we significantly weakens the assumptions imposed on the data matrix.
%Finally, 
%though the sampling scheme in \cite{high-rank-matrix-completion} is more practical than ours (does not need sampling entire rows),
However, \cite{high-rank-matrix-completion}
requires an exponential number of data points to effectively recover the underlying subspaces.
In contrast, in our analysis $N$ only needs to scale polynomially with $r$.
In addition, strong distributional assumptions are imposed in \cite{high-rank-matrix-completion} to ensure that 
data points within the same subspace lie close to each other, while our analysis is applicable to the fully general deterministic setting
where no such distributional properties are required.

\section{Dimension reduction methods}\label{sec:dim_reduction}

In this section we review several popular dimensionality reduction methods and show that they are \emph{subspace embeddings}. 
To keep the presentation simple, proofs of results in this section are presented in Appendix \ref{appsec:proof_ose}.
A linear projection $\mat\Psi\in\mathbb R^{p\times d}$ is said to be a subspace embedding if for some $r$-dimensional subspace $\mathcal L\subseteq\mathbb R^d$ the following holds:
\begin{equation}
\Pr_{\mat\Psi}\left[\|\mat\Psi\vct x\|\in (1\pm\epsilon)\|\vct x\|,\forall \vct x\in\mathcal L\right] \geq 1-\delta.
\label{eq_ose}
\end{equation}

The following proposition is a simple property of subspace embeddings.%, which we prove in Appendix \ref{appsec:proof_ose}.
\begin{prop}
Fix $\epsilon,\delta > 0$.
Suppose $\mat\Psi$ is a subspace embedding with respect to a union of subspaces
$\mathcal B=\{\SPAN(\mathcal U^{(\ell)}\cup\mathcal U^{(\ell')});\ell,\ell'\in[k]\}\cup\{\vct x_i, \vct z_i; i\in[N]\}$
with parameters $r'=2r$, $\epsilon'=\epsilon/3$ and $\delta'=2\log((k+N)/\delta)$.
Then with probability $\geq 1-\delta$ for all
$\vct x,\vct y\in\mathcal U^{(\ell)}\cup\mathcal U^{(\ell')}$ we have
\begin{equation}
\big|\langle \vct x,\vct y\rangle - \langle\mat\Psi\vct x,\mat\Psi\vct y\rangle\big| \leq \epsilon\left(\frac{\|\vct x\|^2+\|\vct y\|^2}{2}\right);
\label{eq_well_behave}
\end{equation}
furthermore, for all $\vct x\in\{\vct x_1,\vct z_1,\cdots,\vct x_N,\vct z_N\}$ the following holds:
\begin{equation}
(1-\epsilon)\|\vct x\|_2^2 \leq \|\mat\Psi\vct x\|_2^2 \leq (1+\epsilon)\|\vct x\|_2^2.
\label{eq_well_behave_jl}
\end{equation}
\label{prop_well_behave}
\end{prop}

\paragraph{Random Gaussian projection}
In a random Gaussian projection matrix $\mat\Psi$
each entry $\mat\Psi_{ij}$ is generated from i.i.d. Gaussian distributions $\nml(0,1/\sqrt{p})$,
where $p$ is the target dimension after projection.
Using standard Gaussian tail bounds and Johnson-Lindenstrauss argument we get the following proposition.
%which is proved in Appendix \ref{appsec:proof_ose}.
\begin{prop}
Gaussian random matrices $\mat\Psi\in\mathbb R^{p\times d}$ is a subspace embedding with respect to $\mathcal B$ if
\begin{equation}
%f_1(\epsilon, \log(\delta), r,\log(d))
p\geq 2\epsilon^{-2}(r+\log(2k^2/\delta)
 +\sqrt{4r\log(2k^2/\delta)}
 + 12\log(4N/\delta)).
 \label{eq_wellbehave_jl}
\end{equation}
\label{prop_wellbehave_jl}
\end{prop}

\paragraph{Uniform row sampling}
For uniform row sampling each row in the observed data matrix $\mat X$ is sampled independently at random
so that the resulting matrix has $p$ non-zero rows.
Formally speaking, each row of the projection matrix $\mat\Omega$ is sampled i.i.d. from the distribution
$\Pr\left[\vct\Omega_{i\cdot} = \sqrt{\frac{d}{p}}\vct e_j\right] = \frac{1}{d}$,
where $i\in[p]$, $j\in[d]$ and $\vct e_j$ is a $d$-dimensional indicator vector
with only the $j$th entry not zero.

For uniform row sampling to work, both the observation matrix $\mat X$ and
the column space of the uncorrupted data matrix $\mat Y$ should satisfy certain incoherence conditions.
%That is, for each column the magnitude of the entries should be distributed in a fairly uniform way.
In this paper, we apply the following two types of incoherence/spikiness definitions, which are widely used
in the low rank matrix completion literature \cite{nc-bernstein-2,subspace-detection,adaptive-matrix-completion-arxiv}.

\begin{defn}[Column space incoherence]
Suppose
\footnote{We require both $\vct x_i$ and $\vct z_i$ to be incoherent because the noise vector $\vct z_i$ may not belong to the incoherent subspace $\mathcal U$.}
 $\mathcal U$ is the column space of some matrix and $\mathrm{rank}(\mathcal U)=r$.
Let $\mat U\in\mathbb R^{d\times r}$ be an orthonormal basis of $\mathcal U$.
The incoherence of $\mathcal U$ is defined as
\begin{equation}
\mu(\mathcal U) := \frac{d}{r}\max_{i=1,\cdots,d}{\|\mat U_{(i)}\|_2^2},
\label{eq_mat_incoherence}
\end{equation}
where $\mat U_{(i)}$ indicates the $i$th row of $\mat U$.
\label{defn_mat_incoherence}
\end{defn}

\begin{defn}[Column spikiness]
For a vector $\vct x\in\mathbb R^d$, the spikiness of $\vct x$ is defined as
\begin{equation}
\mu(\vct x) := d\|\vct x\|_{\infty}^2/\|\vct x\|_2^2,
\end{equation}
where $\|\vct x\|_{\infty} = \max_i{|x_i|}$ denotes the vector infinite norm.
\label{defn_vector_incoherence}
\end{defn}
Under these two conditions, uniform row sampling operator $\mat\Omega$ is a subspace embedding.
%We prove in Appendix \ref{appsec:proof_ose}.
\begin{prop}
Suppose $\max_{\ell=1}^k{\mu(\mathcal U^{(\ell)})} \leq \mu_0$ and $\max_{i=1}^N{\max(\mu(\vct x_i),\mu(\vct z_i))} \leq \mu_0$
for some constant $\mu_0>0$.
The uniform sampling operator $\mat\Omega$ is a subspace embedding with respect to $\mathcal B$ if
\begin{equation}
%f_2(\epsilon,\log(\delta),r,\log(d)) \\
p \geq 8\epsilon^{-2}\mu_0(r\log(4rk^2/\delta) + \log(8N/\delta)).
\label{eq_wellbehave_unisample}
\end{equation}
\label{prop_wellbehave_unisample}
\end{prop}

\paragraph{FJLT and sketching}
The Fast Johnson-Lindenstrauss Transform (FJLT, \citealp{fjlt}) computes a compressed version of a data matrix $\mat X\in\mathbb R^{d\times N}$
using $O(d\log d+p)$ operations instead of $O(pd)$ per column with high probability.
%In contrast,  general dense projectors require $O(Ndp)$ operations to compute the compressed matrix.
The projection matrix $\mat\Phi$ can be written as $\mat\Phi=\mat P\mat H\mat D$,
where $\mat P\in\mathbb R^{p\times d}$ is a sparse JL matrix, $\mat H\in\mathbb R^{d\times d}$ is a deterministic Walsh-Hadamard matrix
and $\mat D\in\mathbb R^{d\times d}$ is a random diagonal matrix.
Details of FJLT can be found in \cite{fjlt}.
%For detailed construction of FJLT we refer the readers to the seminal work by Ailon and Chazelle \cite{fjlt}.

Sketching \citep{count-sketch,cs-ose} is another powerful tool for dimensionality reduction on sparse inputs.
The sketching operator $\mat S:\mathbb R^d\to\mathbb R^p$ is constructed as $\mat S=\mat\Pi\mat\Sigma$,
where $\mat\Pi$ is a random permutation matrix and $\mat\Sigma$ is a random sign diagonal matrix.
The projected vector $\mat S\vct x$ can be computed in $O(\mathrm{nnz}(\vct x))$ time,
where $\mathrm{nnz}(\vct x)$ is the number of nonzero entries in $\vct x$.

%Proposition \ref{prop_wellbehave_fjlt} shows that FJLT is well-behaved.
%Nevertheless, the sample complexity in Eq. (\ref{eq_wellbehave_fjlt}) is $O(r)$ worse than the sample complexity in Eq. (\ref{eq_wellbehave_jl})
%for standard Johnson-Lindenstrauss transform.
%This is because we require \emph{uniform} inner product preservation within a low rank subspace,
%which is stronger than the original guarantee obtained in \cite{fjlt}.
%The increased sample complexity can be viewed as an extra price paid for fast compressing operators.

The following two propositions show that both FJLT and sketching are subspace embeddings,
meaning that with high probability the inner product and norm of any two vectors on a low-dimensional subspace are preserved uniformly.
In fact, they are \emph{oblivious} in the sense that they work for any low-dimensional subspace $\mathcal L$.
\begin{prop}[{\citealp{cs-ose}}]
The FJLT operator $\mat\Phi$ is an oblivious subspace embedding if $p=\Omega(r/\epsilon^2)$,
with $\delta$ considered as a constant.
\label{prop_wellbehave_fjlt}
\end{prop}
\begin{prop}[{\citealp{tensor-sketch-ose}}]
The sketching operator $\mat S$ is an oblivious subspace embedding if $p=\Omega(r^2/(\epsilon^2\delta))$.
\label{prop_wellbehave_sketching}
\end{prop}

\subsection{Simulations}

\begin{figure}[t]
\centering
\includegraphics[width=6cm]{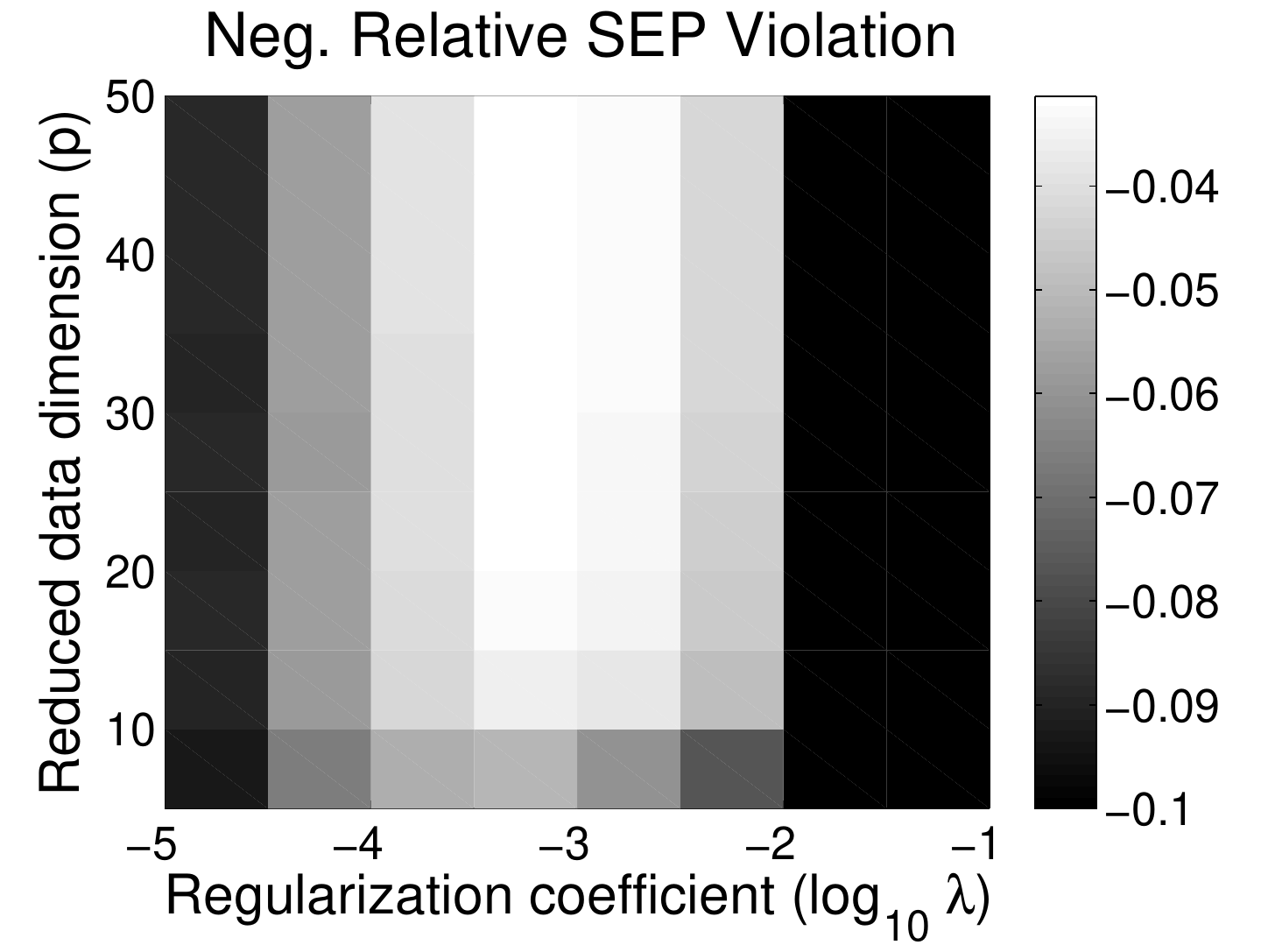}
\includegraphics[width=6cm]{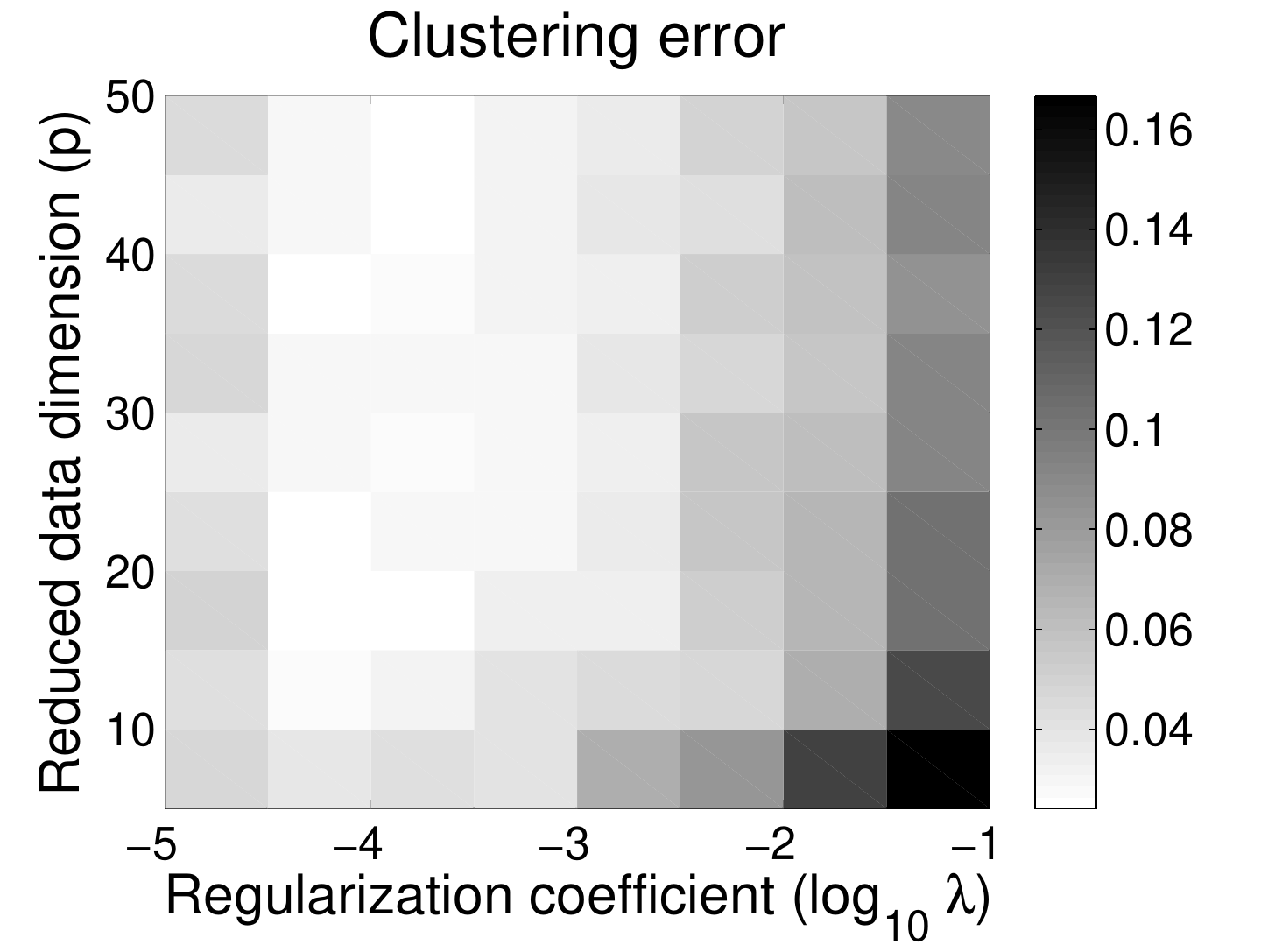}
\caption{Relative SEP violation (left) and clustering error for Lasso SSC on the Hopkins-155 dataset.
The rightmost two columns in the left figure indicate trivial solutions.
White indicates good similarity graph or clustering and black indicates poor similarity graph or clustering.}
\label{fig_hopkins_ssc}
\end{figure}

\begin{figure}[t]
\centering
\includegraphics[width=6cm]{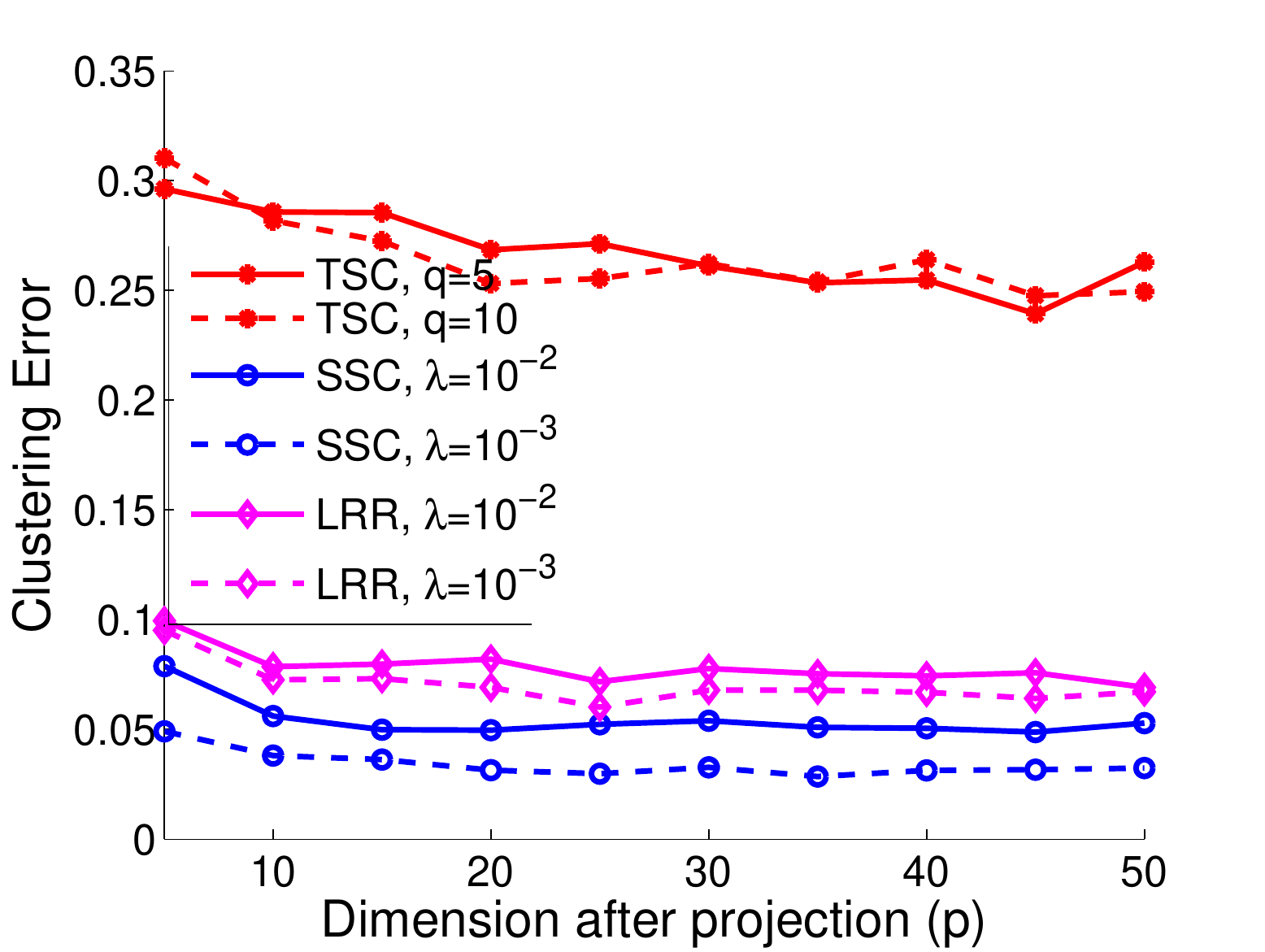}
\includegraphics[width=6cm]{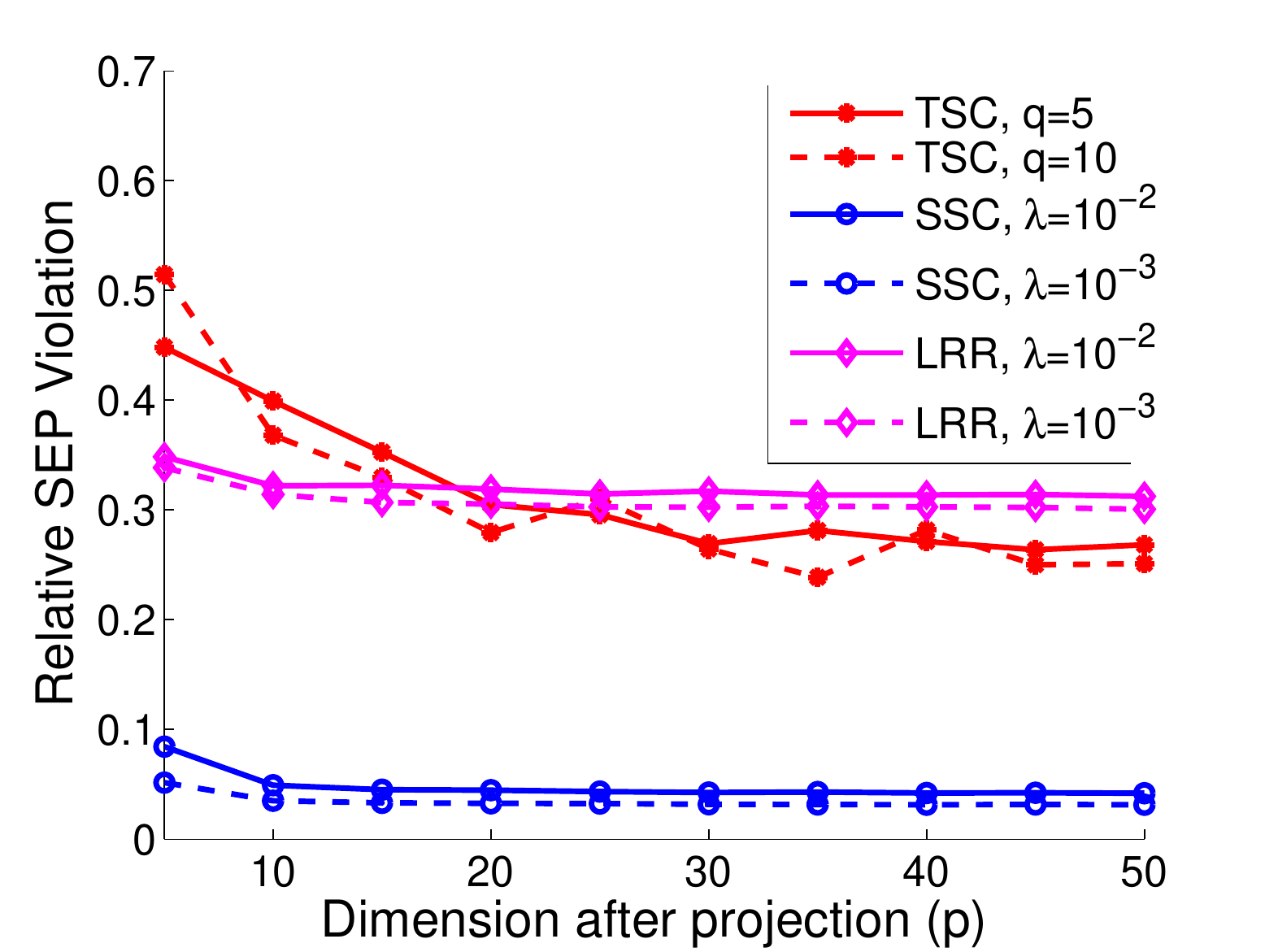}
\caption{Comparison of clustering error (left) and relative SEP violation (right) for Lasso SSC, TSC and LRR on the Hopkins-155 dataset.}
\label{fig_hopkins_compare}
\end{figure}

To gain some intuition into the performance of the SSC algorithm on dimensionality-reduced data,
we report empirical results on \emph{Hopkins-155}, a motion segmentation data set that is specifically designed to serve as a benchmark for subspace clustering algorithms \citep{hopkins155}.
The ambient dimension $d$ in the data set ranges from 112 to 240, and we compress the data points into $p$ dimension using random Gaussian projection,
with $p$ taking the values from 5 to 50.
We compare Lasso SSC with TSC \citep{tsc} and LRR \citep{lrr}.
The Lasso SSC algorithm is implemented using augmented Lagrangian method (ALM, \cite{bertsekas2014constrained}).
The LRR implementation is obtained from \cite{lrr-software}.
All algorithms are implemented in Matlab.

For evaluation, we report both clustering error and the relative violation of the Self-Expressiveness Property (SEP).
Clustering error is defined as the percentage of mis-clustered data points up to permutation.
The relative violation of SEP characterizes how much the obtained similarity matrix $\mat C$ violates the self-expressiveness property.
It was introduced in \cite{noisy-ssc} and defined as
\begin{equation}
\text{RelViolation}(\mat C,\mathcal M) = \frac{\sum_{(i,j)\notin\mathcal M}{|\mat C|_{ij}}}{\sum_{(i,j)\in\mathcal M}{|\mat C|_{ij}}},
\end{equation}
where $(i,j)\in\mathcal M$ means $\vct x_i$ and $\vct x_j$
belong to the same cluster and vice versa.

Figure \ref{fig_hopkins_ssc} shows that the relative SEP violation of Lasso SSC goes up when the projection dimension $p$ decreases,
or the regularization parameter $\lambda$ is too large or too small.
The clustering accuracy acts accordingly.
%with the exception of very large $\lambda$ values under which we get very sparse self-regression vectors
%and hence connectivity of the similarity graph is affected.
In addition, in Figure \ref{fig_hopkins_compare} we report the clustering error and relative SEP violation for Lasso SSC, TSC and LRR on Hopkins-155.
Both clustering error and relative SEP violation are averaged across all 155 sequences.
Figure \ref{fig_hopkins_compare} also indicates that Lasso SSC outperforms TSC and LRR under various regularization and projection dimension settings,
which is consistent with previous experimental results \citep{ssc}.

\section{Main results}\label{sec:main_result}
%\vspace*{-0.15cm}

%We present general geometric separation conditions for Lasso sparse subspace clustering (Eq.~(\ref{eq_lasso_ssc}))
%to succeed for dimensionality-reduced data in the fully deterministic setting; that is, both subspaces and data points within subspaces are deterministically distributed.
%In addition, our analysis reveals that SSC is able to robustly detect the correct subspaces with substantially compressed data even when the data points are adversarially perturbed, stochastically contaminated, or subject to formal privacy constraints. These contributions significantly expand the previous provable results on the same subject that works only with noiseless data generated from the ``semi-random'' model \citep{sc-dr-semirandom}.
%

We present general geometric success conditions for Lasso SSC on dimensionality-reduced data, under the fully deterministic setting where both the underlying low-dimensional subspaces
and the data points on each subspace are placed deterministically.
We first describe the result for the noiseless case and then the results are extended to handle a small amount of adversarial perturbation or a much larger amount of stochastic noise. 
In addition, implications of our success conditions under the much stronger semi-random and fully random models are discussed.
%A performance guarantee under differential privacy can then be stated as a simple corollary of the noisy recovery result.
%for both noiseless and noisy settings.
The basic idea common in all of the upcoming results is to show that the subspace incoherence and inradius
\footnote{Both subspace incoherence and inradius are key quantities appearing in analysis of sparse subspace clustering algorithms and will be defined later.}
 (therefore the geometric gap) are approximately preserved under dimension reduction.

%bound the perturbation of subspace incoherence and inradius separately after dimension reduction.

%is to bound the perturbation of subspace incoherence and inradius separately after dimension reduction.

%Note that Lasso SSC with $\lambda > 0$ is required even for noiseless inputs
%because the a positive $\lambda$ makes the dual optimization problem in Eq. (\ref{eq_lasso_dual}) strongly convex
%and hence perturbation bounds on dual directions can be proved by bounding the perturbation of objective values.

\subsection{Deterministic model: the noiseless case}

We consider first the noiseless case where $\mat Z=\mat 0$.
We begin our analysis with two key concepts introduced in the seminal work of \cite{sc-geometric}:
\emph{subspace incoherence} and \emph{inradius}.
Subspace incoherence characterizes how well the subspaces associated with different clusters are separated.
It is based on the \emph{dual direction} of the optimization problem in Eq. (\ref{eq_ssc}) and (\ref{eq_lasso_ssc}), which is defined as follows:
\begin{defn}[Dual direction, \citealp{sc-geometric,noisy-ssc}]
Fix a column $\vct x$ of $\mat X$ belonging to subspace $\mathcal U^{(\ell)}$.
Its dual direction $\vct\nu(\vct x)$ is defined as the solution to the following dual optimization problem:
\footnote{For exact SSC simply set $\lambda = 0$.}
\begin{eqnarray}
\max_{\vct\nu\in\mathbb R^d}{\langle\vct x,\vct\nu\rangle - \frac{\lambda}{2}\vct\nu^\top\vct\nu},\quad
\text{s.t. } \|\mat X^\top\vct\nu\|_{\infty} \leq 1.
\label{eq_lasso_dual}
\end{eqnarray}
Note that Eq. (\ref{eq_lasso_dual}) has a unique solution when $\lambda > 0$.
\label{defn_dual_direction}
\end{defn}

The subspace incoherence for $\mathcal U^{(\ell)}$, $\mu_\ell$,
is defined in Eq.~(\ref{eq_subspace_incoherence}).
Note that it is not related to the column subspace incoherence defined in Eq. (\ref{eq_mat_incoherence}).
The smaller $\mu_\ell$ is the further $\mathcal U^{(\ell)}$ is separated from the other subspaces.
\begin{defn}[Subspace incoherence, \cite{sc-geometric,noisy-ssc}]
Subspace incoherence $\mu_\ell$ for subspace $\mathcal U^{(\ell)}$ is defined as
\begin{equation}
\mu_\ell := \max_{\vct x\in\mat X\backslash\mat X^{(\ell)}}{\|\mat V^{(\ell)\top}\vct x\|_{\infty}},
\label{eq_subspace_incoherence}
\end{equation}
where $\mat V^{(\ell)} = (\vct v(\vct x_1^{(\ell)}),\cdots,\vct v(\vct x_{N_\ell}^{(\ell)}))$ and
$\vct v(\vct x) = \mathcal P_{\mathcal U^{(\ell)}}\vct\nu(\vct x)/{\|\mathcal P_{\mathcal U^{(\ell)}}\vct\nu(\vct x)\|_2}$.
$\vct\nu(\vct x)$ is the dual direction of $\vct x$ defined in Eq. (\ref{eq_lasso_dual})
and $\mathcal U^{(\ell)}$ is the low-dimensional subspace on which $\vct x$ lies.
\label{defn_subspace_incoherence}
\end{defn}

The concept of inradius characterizes how well data points are distributed within a single subspace.
More specifically, we have the following definition:
\begin{defn}[Inradius, \citealp{sc-geometric}]
For subspace $\mathcal U^{(\ell)}$, its inradius $\rho_\ell$ is defined as
\begin{equation}
\rho_\ell := \min_{i=1,\cdots,N_\ell}{r(\mathcal Q(\mat X_{-i}^{(\ell)}))},
\label{eq_rho}
\end{equation}
where $r(\cdot)$ denotes the radius of the largest ball inscribed in a convex body.
\label{defn_inradius}
\end{defn}
The larger $\rho_\ell$ is, the more uniformly data points are distributed in the $\ell$th subspace.
%Note that unlike subspace incoherence, the inradius is defined in terms of the uncorrupted data $\mat Y$.
We also remark that both $\mu_\ell$ and $\rho_\ell$ are between 0 and 1 because of normalization.

With the characterization of subspace incoherence and inradius, we are now ready to present our main result,
which states sufficient success condition for Lasso SSC on dimensionality-reduced noiseless data under the fully deterministic setting.
\begin{thm}[Compressed SSC on noiseless data]
Suppose $\mat X\in\mathbb R^{d\times N}$ is a noiseless input matrix
with subspace incoherence $\{\mu_\ell\}_{\ell=1}^k$ and inradii $\{\rho_\ell\}_{\ell=1}^k$.
Assume $\mu_\ell < \rho_\ell$ for all $\ell\in\{1,\cdots,k\}$.
Let $\widetilde{\mat X}'$ be the normalized data matrix after compression.
Assume $0<\lambda< \rho/2$, where $\rho=\min_{\ell}\rho_\ell$.
If $\mat\Psi$ satisfies Eq. (\ref{eq_well_behave},\ref{eq_well_behave_jl}) with parameter $\epsilon$
then Lasso SSC satisfies subspace detection property %with probability $\geq 1-\delta$,
if $\epsilon$ is upper bounded by
\begin{equation}
\epsilon = O\left(\Delta^2 \lambda^{3/2}\right),
\end{equation}
%where $c_1 > 0$ is some absolute constant and
where $\Delta = \min_{\ell}{(\rho_{\ell}-\mu_{\ell})}$ is the minimum gap between subspace incoherence and inradius for each subspace.
\label{thm_main_noiseless}
\end{thm}

We make several remarks on Theorem \ref{thm_main_noiseless}.
First, an upper bound on $\epsilon$ implies a lower bound about projection dimension $p$,
and exact $p$ values vary for different data compression schemes.
For example, if Gaussian random projection is used then $\epsilon=O(\Delta^2\lambda^{3/2})$ is satisfied with $p=\Omega(\Delta^{-2}\lambda^3r\log N)$.
%The major difference between our result and previous analysis on exact subspace clustering
%is that the regularization coefficient $\lambda$ cannot be too small
In addition, even for noiseless data the regularization coefficient $\lambda$ cannot be too small if projection error $\epsilon$ is present
(recall that $\lambda\to 0$ corresponds to the exact SSC formulation).
This is because when $\lambda$ goes to zero the strong convexity of the dual optimization problem decreases.
As a result, small perturbation on $\mat X$ could result in drastic changes of the dual direction
and Lemma \ref{lem_subspace_incoherence_noiseless} fails subsequently.
On the other hand, as $\lambda$ increases the similarity graph connectivity decreases
because the optimal solution to Eq. (\ref{eq_lasso_ssc}) becomes sparser.
To guarantee the obtained solution is nontrivial (i.e., at least one nonzero entry in $\vct c_i$),
$\lambda$ must not exceed $\rho/2$.

\subsection{Deterministic model: the noisy case}

When data are corrupted by either adversarial or stochastic noise, 
we can still hope to get success conditions for Lasso SSC provided that the magnitude of noise is upper bounded.
The success conditions can again be stated using the concepts of subspace incoherence and inradius.
Although the definition of subspace incoherence in Eq. (\ref{eq_subspace_incoherence}) remains unchanged (i.e., defined in terms of the noisy data $\mat X$),
the definition of inradius needs to be slightly adjusted under the noisy setting as follows:
\begin{defn}[Inradius for noisy SSC, \cite{noisy-ssc}]
Let $\mat X=\mat Y+\mat Z$ where $\mat Y$ is the uncorrupted data, $\mat Z$ is the noise matrix and $\mat X$ is the observation matrix.
For subspace $\mathcal U^{(\ell)}$, its inradius $\rho_\ell$ is defined as
\begin{equation}
\rho_\ell := \min_{i=1,\cdots,N_\ell}{r(\mathcal Q(\mat Y_{-i}^{(\ell)}))},
\label{eq_rho}
\end{equation}
where $r(\cdot)$ denotes the radius of the largest ball inscribed in a convex body.
\label{defn_inradius_noisy}
\end{defn}
As a remark, under the noiseless setting we have $\mat X=\mat Y$ and Definition \ref{defn_inradius_noisy} reduces to the definition of inradius for noiseless data.

Finally, we have Theorem~\ref{thm_main_noisy} and Theorem~\ref{thm_main_noisy_rand}.
%as simple consequences of Corollary~\ref{cor_perturbation_mu_noisy} and Lemma~\ref{lem_inradius_preservation}.
\begin{thm}[Compressed-SSC under deterministic noise]
Suppose $\mat X=\mat Y+\mat Z$ is a noisy input matrix
with subspace incoherence $\{\mu_\ell\}_{\ell=1}^k$ and inradii $\{\rho_\ell\}_{\ell=1}^k$.
Assume $\max_i{\|\vct z_i\|_2} \leq\eta$ and $\mu_\ell < \rho_\ell$ for all $\ell\in\{1,\cdots,k\}$.
%Assume $\mat Z_{ij}\sim\nml(0,\frac{\sigma^2}{d})$ for some noise magnitude parameter $\sigma$.
Suppose $\widetilde{\mat X}' = \widetilde{\mat Y}' + \widetilde{\mat Z}$
where $\widetilde{\mat Y}'$ is the normalized uncorrupted data matrix after compression
and $\widetilde{\mat Z}=\mat\Psi\mat Z$ is the projected noise matrix.
If $\mat\Psi$ satisfies Eq. (\ref{eq_well_behave},\ref{eq_well_behave_jl}) with parameter $\epsilon$ and $\lambda$ satisfies %$\lambda = \rho/4$,
\begin{equation}
\frac{24\eta}{\Delta} < \lambda < \rho(1-\epsilon)-6\eta,
\label{eq_lambda_detailed_range}
\end{equation}
then Lasso SSC satisfies the subspace detection property if the approximation error $\epsilon$ and noise magnitude $\eta$ satisfy
\begin{equation}
\max(\epsilon,\eta) = O\left(\Delta^2 \lambda^{3/2}\right).
\label{eq_eps_noisy}
\end{equation}
%where $c_2 > 0$ is some absolute constant and
Here $\Delta = \min_\ell{(\rho_\ell-\mu_\ell)}$ is the minimum gap between subspace incoherence and inradius.
\label{thm_main_noisy}
\end{thm}

\begin{thm}[Compressed-SSC under stochastic noise]
Suppose $\mat X=\mat Y+\mat Z$ is a noisy input matrix
with subspace incoherence $\{\mu_\ell\}_{\ell=1}^k$ and inradii $\{\rho_\ell\}_{\ell=1}^k$.
%Assume $\max_i{\|\vct z_i\|_2} \leq\eta$ and $\mu_\ell < \rho_\ell$ for all $\ell\in\{1,\cdots,k\}$.
Assume $\mat Z_{ij}\sim\nml(0,\frac{\sigma^2}{d})$ for some noise magnitude parameter $\sigma$.
Suppose $\widetilde{\mat X}' = \widetilde{\mat Y}' + \widetilde{\mat Z}$
where $\widetilde{\mat Y}'$ is the normalized uncorrupted data matrix after compression
and $\widetilde{\mat Z}=\mat\Psi\mat Z$ is the projected noise matrix.
If the spectral norm of $\mat\Psi\in\mathbb R^{p\times d}$ is upper bounded by $\|\mat\Psi\|_2\leq\psi\sqrt{\frac{d}{p}}$ for some constant $\psi$ (i.e., $\psi=O(1)$)
and in addition
$\mat\Psi$ satisfies Eq. (\ref{eq_well_behave},\ref{eq_well_behave_jl}) with parameter $\epsilon$ and $\lambda$ satisfies
\begin{equation}
C_1\cdot \frac{\sigma(1+\sigma)}{\Delta}\sqrt{\frac{\log N}{d}} < \lambda < \frac{\rho}{2}-C_2\cdot \sigma(1+\sigma)\sqrt{\frac{\log N}{d}}
\label{eq_lambda_detailed_range_stochastic}
\end{equation}
for some universal constants $C_1,C_2 > 0$,
then Lasso SSC satisfies the subspace detection property if the approximation error $\epsilon$ and noise magnitude $\sigma$ satisfy
\begin{equation}
\epsilon=O\left(\Delta^2\lambda^{3/2}\right), \quad
\sigma = O\left(\min\left\{ \frac{\Delta^{1/2}\lambda^{1/2}d^{1/4}}{\log^{1/4}N}, \frac{\Delta\rho^{1/2}d^{1/4}}{r^{1/4}\log^{5/4} N}, \frac{\Delta^2\lambda^{3/2}d^{1/2}}{r^{1/2}\log^{3/2}N} \right\}\right),
\label{eq_main_noisy_rand}
\end{equation}
%where $c_2 > 0$ is some absolute constant and
where $\Delta = \min_\ell{(\rho_\ell-\mu_\ell)}$ is the minimum gap between subspace incoherence and inradius.
\label{thm_main_noisy_rand}
\end{thm}

Before we proceed some clarification on the ambient data dimension $d$ is needed.
Eq.~(\ref{eq_main_noisy_rand}) seems to suggest that the noise variance $\sigma$ \emph{increases} with $d$.
While this is true if noise magnitude is measured in terms of the $\ell_2$ norm $\|\vct z\|_2$,
we remark that \emph{coordinate-wise} noise variance does not increase with $d$, because each one of the $d$ coordinates of $z$
is a Gaussian random variable with variance $\sigma^2/d$.

These results put forward an interesting view of the subspace clustering problem in terms of resource allocation. The critical geometric gap $\Delta$ (called ``Margin of Error'' in \citet{noisy-ssc}) can be viewed as the amount of resource that we have for a problem while preserving the subspace detection property.
It can be used to tolerate noise, compress the data matrix, or alleviate the graph connectivity problem of SSC \cite{wang2013provable}.
With larger geometric gap $\Delta$, the approximation error $\epsilon$ from dimensionality reduction can be tolerated at a larger level,
which implies that the original data $\mat X$ can be compressed more aggressively, at a smaller dimension $p$ after compression, without losing the SDP property
of sparse subspace clustering.
The results also demonstrate trade-offs between noise tolerance and dimensionality reduction.
For example, at a higher level of noise variance $\sigma$ or $\eta$ the regularization parameter $\lambda$ must be set at a lower level,
according to conditions in Eqs.~(\ref{eq_lambda_detailed_range},\ref{eq_lambda_detailed_range_stochastic});
subsequently, the approximation error $\epsilon$ in dimensionality reduction needs to be smaller to ensure success of SSC on the projected data,
which places a higher lower bound on the dimension $p$ one can compress the original data into.
%For example, if the noise level is high then it will use more of $\Delta$ and as a result we can only compress the data less aggressively,
%as shown in Eq. (\ref{eq_eps_noisy}) and (\ref{eq_main_noisy_rand}).
%If we compress the data for computational benefits or privacy considerations, part of this resource gets used, and then whatever that is left in the ``Margin of Error'' can be used to tolerate a small amount of arbitrary noise, or larger amount of stochastic noise. The more aggressive we compress the data, the less reliable the method is to noise and model uncertainty. We note that a similar observation is made in \citet{wang2013provable}, that the ``Margin of Error'' can be used to alleviate the graph connectivity problem of SSC.

\subsection{Semi-random and Fully-random models}

In this section we consider random \emph{data} models, where either the data points or the underlying low-dimensional subspaces are assumed to be drawn i.i.d.~from a uniform distribution.
Under the semi-random model the underlying subspaces $\mathcal U^{(\ell)}$, $\ell=1,\cdots,k$ are still assumed to be fixed but unknown.
However, we place stochastic conditions on the data points by assuming that each data point is drawn uniformly at random from the unit sphere of the corresponding low-dimensional subspace.
This assumption makes the success conditions of Lasso SSC more transparent, as the success conditions now only depend on the number of data points per subspace
and the \emph{affinity} between different subspaces, which is formally defined as follows:
\begin{defn}[Subspace affinity (normalized), \citep{sc-geometric,noisy-ssc}]
For two subspaces $\mathcal U^{(\ell)},\mathcal U^{(\ell')}$ of intrinsic dimension $r_\ell$ and $r_{\ell'}$,
the \emph{affinity} between $\mathcal U^{(\ell)}$ and $\mathcal U^{(\ell')}$ is defined as
\begin{equation}
\affn(\mathcal U^{(\ell)},\mathcal U^{(\ell')}) = \sqrt{\frac{\cos^2(\theta_1) + \cdots + \cos^2(\theta_{r_\ell\wedge r_{\ell'}})}{r_\ell r_{\ell'}}},
\end{equation}
where $\theta_1,\cdots,\theta_{r_\ell\wedge r_{\ell'}}$ are canonical angles between $\mathcal U^{(\ell)}$ and $\mathcal U^{(\ell')}$.
Note that $\affn(\mathcal U^{(\ell)},\mathcal U^{(\ell')})$ is always between 0 and 1, with smaller value indicating one subspace being further apart from the other.
\end{defn}

We are now able to state our main theorem on success conditions of dimensionality-reduced noisy SSC under the semi-random model.
\begin{thm}[Compressed SSC under Semi-random Model]
Suppose $\mat X=\mat Y+\mat Z$ is a noisy input matrix with $\mat Y^{(\ell)}$ sampled uniformly at random from the unit sphere in $\mathcal U^{(\ell)}$
and $\mat Z_{ij}\sim\nml(0,\frac{\sigma^2}{d})$ for some noise magnitude parameter $\sigma$.
Assume in addition that $N_\ell\geq \kappa r_\ell$ for some $\kappa > 1$ and the subspace affinity $\{\affn(\mathcal U^{(\ell)},\mathcal U^{(\ell')})\}_{\ell,\ell'=1}^k$ satisfies
\begin{equation}
\min_{\ell'\neq\ell}\affn(\mathcal U^{(\ell)}, \mathcal U^{(\ell')}) = O\left(\frac{1}{\log^2(kN)}\sqrt{\frac{\log\kappa}{r}}\right),\;\;\;\;
\forall \ell=1,\cdots,k.
\label{eq_semirandom_affinity}
\end{equation}
Suppose $\widetilde{\mat X}' = \widetilde{\mat Y}' + \widetilde{\mat Z}$
where $\widetilde{\mat Y}'$ is the normalized uncorrupted data matrix after compression
and $\widetilde{\mat Z}=\mat\Psi\mat Z$ is the projected noise matrix.
If the spectral norm of $\mat\Psi\in\mathbb R^{p\times d}$ is upper bounded by $\|\mat\Psi\|_2\leq\psi\sqrt{\frac{d}{p}}$ for some constant $\psi$ (i.e., $\psi=O(1)$)
and in addition
$\mat\Psi$ satisfies Eq. (\ref{eq_well_behave},\ref{eq_well_behave_jl}) with parameter $\epsilon$ and $\lambda \asymp \sqrt{\log\kappa/r}$,
then Lasso SSC satisfies the subspace detection property with probability $\geq 1-O(1/N)$ if the approximation error $\epsilon$ and noise magnitude $\sigma$ satisfy
\begin{equation}
\epsilon = O\left(\frac{\log^{7/4}\kappa}{r^{7/4}}\right), \;\;\;\;
\sigma = O\left(\min\left\{ \frac{d^{1/4}\log^{3/4}\kappa}{r\log^{5/4} N}, \frac{d^{1/2}\log^{7/4}\kappa }{r^{9/4}\log^{3/2}N} \right\}\right).
\label{eq_main_noisy_semirandom}
\end{equation}
Furthermore, as a corollary, if $\mat\Psi\in\mathbb R^{p\times d}$ is the random Gaussian projection matrix (i.e., $\mat\Psi_{ij}\sim\nml(0,1/p)$) with
$p$ the dimension after projection,
then Lasso SSC satisfies the subspace detection property with probability $\geq 1-O(1/N)$ if the noise magnitude $\sigma$ satisfies the constraint as in Eq. (\ref{eq_main_noisy_semirandom})
and the projected dimension $p$ is at least
\begin{equation}
p = \Omega\left(r^{9/2}\cdot \frac{\log(kN)}{\log^{7/2}\kappa}\right).
\label{eq_main_noisy_semirandom_p}
\end{equation}
\label{thm_main_semirandom}
\end{thm}

As a remark, Theorem \ref{thm_main_semirandom} shows that using the Gaussian random projection operator,
the dimension after projection $p$ depends polynomially on the intrinsic dimension $r$.
Such dependency is unavoidable, as one cannot hope to compress the data to the point that data dimension after compression is smaller than the intrinsic dimension.
On the other hand, our analysis shows that $p$ depends only \emph{poly-logarithmically} on both the number of subspaces $k$ and the number of data points $N$.
Such dependency is significantly better than treating the entire union-of-subspace model as an agglomerate low-rank model,
which would then require $\Omega(kr\log N)$ dimension after compression.
However, we conjecture the $r^{9/2}$ term in Eq. (\ref{eq_main_noisy_semirandom_p}) is loose (which comes up in our argument of strong convexity of the dual problem of SSC)
and could be improved to an even lower order of polynomial function.
Finally, we note that $p$ needs to increase with $N$, the total number of data points, which might seem counter-intuitive.
This is because success of the SSC algorithm is defined in terms of self-expressiveness property over \emph{all} data points,
which becomes more difficult to satisfy with more data points.

We next turn to the fully-random model, where not only the data points but also the underlying subspaces $\mathcal U^{(1)}, \cdots,\mathcal U^{(k)}$ are 
assumed to be drawn i.i.d.~uniformly at random.
In this even simpler model, we have the following theorem that directly follows from Theorem \ref{thm_main_semirandom}:
\begin{thm}[Compressed SSC under fully random model]
With the same notation and conditions in Theorem \ref{thm_main_semirandom},
except that the condition on subspace affinity, Eq. (\ref{eq_semirandom_affinity}),
is replaced by the following condition that involves only the number of data points $N$, the ambient dimension $d$ and the data ratio $\kappa$:
\begin{equation}
r = O\left(\frac{d\log\kappa}{\log N}\right).
\end{equation}
Then under the random Gaussian projection operator,
Lasso SSC succeeds with probability $\geq 1-O(1/N)$ if $\lambda\asymp\sqrt{\log(\kappa)/r}$
and the number of dimension after compression $p$ satisfies
\begin{equation}
p = \Omega\left(r^{9/2}\cdot\frac{\log(kN)}{\log^{7/2}\kappa}\right).
\label{eq_fully_random_p}
\end{equation}
\label{thm_main_fullyrandom}
\end{thm}

\section{Application to privacy preserving subspace clustering}

Dimensionality reduction is useful in many practical applications involving subspace clustering, as explained in the introduction.
In this section, we discuss one particular motivation of compressing data before data analysis in order to protect data privacy.
The privacy issue of subspace clustering has received research attention recently \citep{private-ssc},
as it is applied to sensitive data sets such as medical/genetic or movie recommendation data \citep{sc-movie-rating,sc-disease}.
Nevertheless, there has been no prior work on formally establishing privacy claims for sparse subspace clustering algorithms.

In this section we investigate differentially private sparse subspace clustering under a random projection dimension reduction model. This form of privacy protection is called ``matrix masking'' and has a long history in statistical privacy and disclosure control (see, e.g., \citealp{duncan1991enhancing,willenborg1996statistical,hundepool2012statistical}).
It has been formally shown more recently that random projections (at least with Gaussian random matrices) protect information privacy \citep{zhou2009compressed}.
Stronger privacy protection can be enforced by injecting additional noise to the dimension reduced data \citep{data-privacy-jl}. 
Algorithmically, this basically involves adding iid Gaussian noise to the data after we apply a Johnson-Lindenstrauss transform $\mat\Psi$ of choice to $\mat X$ and normalize every column. 
This procedure guarantees differential privacy \citep{dwork2006calibrating,dwork2006differential} at the attribute level, which prevents any single entry of the data matrix from being identified ``for sure'' given the privatized data and arbitrary side information.
The amount of noise to add is calibrated according to how ``unsure'' we need to be and how ``spiky'' (Definition~\ref{defn_vector_incoherence}) each data point can be.
We show in Sec.~\ref{sec:privacy} and \ref{sec:utility} that the proposed variant of SSC algorithm achieves perfect clustering with high probability while subject to formal privacy constraints.
We also prove in Sec. \ref{sec:impossibility} that a stronger user-level privacy constraint cannot be attained when perfect clustering of each data point is required.
\citet{private-ssc} discussed alternative solutions to this dilemma by weakening the utility claims.

\subsection{Privacy claims}\label{sec:privacy}

In classic statistical privacy literature, transforming data set $\mat X$ by taking $\tilde{\mat X} = \mat A\mat X  + \mat E$ for some random matrix $\mat A$ and $\mat E$ is called \emph{matrix masking}. \cite{zhou2009compressed} show that random compression allows the mutual information of the output $\tilde{\mat X}$ and raw data $\mat X$ to converge to $0$ with rate $O(p/d)$ even when $\mat M=0$, and their result directly applies to our problem. The guarantee suggests that the amount of information in the compressed output $\tilde{\mat X}$ about the raw data $\mat X$ goes to 0 as the ambient dimension $d$ gets large.

On the other hand, if $\mat E\neq\mat 0$ is an iid Gaussian noise matrix,
we can protect the $(\varepsilon,\delta)$-differential privacy of every data entry.
More specifically, we view the matrix $\mat X$ as a data collection of $N$ users,
each corresponding to a column in $\mat X$ associated with a $d$-dimensional attribute vector.
Each entry in $\mat X$ then corresponds to a particular attribute of a specific user.
The formal definition of attribute differential privacy notion is given below:
\begin{defn}[Attribute Differential Privacy]\label{defn:attribute-privacy}
Suppose $\cO$ is the set for all possible outcomes.
We say a randomized algorithm $\cA: \R^{d\times N} \rightarrow \cO $ is $(\varepsilon,\delta)$-differentially private at attribute level if
$$\Pr(\cA(\mat X)\in \cS) \leq e^{\varepsilon}\Pr(\cA(\mat X')\in \cS) + \delta$$
for any measurable outcome $\cS\subset \cO$, any $\mat X$ and $\mat X'$ that differs in only one entry.
\end{defn}
This is a well-studied setting in \citep{data-privacy-jl}. It is weaker than protecting the privacy of individual users (i.e., entire columns in $\mat X$),
 but much stronger than the average protection via mutual information. In fact, it forbids any feature of an individual user from being identified ``for sure'' by an adversary with arbitrary side information.
\begin{thm}\label{thm:privacy}
Assume the data (and all other users that we need to protect) satisfy column spikiness conditions with parameter $\mu_0$
 as in Definition~\ref{defn_vector_incoherence}.
 Let $\mat\Psi$ be a Johnson-Lindenstrauss transform with parameter $\epsilon$.
Releasing compressed data $\widetilde{\mat X}' = \mathrm{Normalize}(\mat\Psi \mat X) + \cN(0,\sigma^2/d\cdot \mat I_{p\times d})$ with $\sigma = \frac{1+\epsilon}{1-\epsilon} \sqrt{\frac{32\mu_0\log(1.25/\delta)}{\varepsilon^2}}$ preserves attribute-level $(\varepsilon,\delta)$-differential privacy.
\end{thm}
Note that in Theorem \ref{thm:privacy} the exact value of $\mu_0$ is not necessary: an upper bound of $\mu_0$ would be sufficient,
which results in more conservative differentially private procedures.

The proof involves working out the $\ell_2$-sensitivity of the operator $\text{Normalize}(\mat\Psi(\cdot))$ in terms of column incoherence $\mu_0$ and apply ``Gaussian Mechanism'' \citep{dwork2006differential,dwork2013algorithmic}.
We defer the proof to Sec.~\ref{subsec:proof-privacy}.
Note that differential privacy is close to ``post-processing'', meaning that any procedure on the \emph{released} data does not change
the privacy level.
Therefore, applying SSC on the released data $\widetilde{\mat X}'$ injected with noise remains a $(\epsilon,\delta)$-differentially private procedure.
%By the closeness to post-processing property of differential privacy, the subsequent subspace clustering results protects the same level of privacy. Details are given as follows.

\subsection{Utility claims}\label{sec:utility}

We show that if column spikiness $\mu_0$ is a constant, Lasso-SSC is able to provably detect the correct subspace structures, despite privacy constraints.
 \begin{cor}\label{cor:utility}
  Let the raw data $\mat X$ be compressed and privatized using the above described mechanism to get $\widetilde{\mat X}$.
  Assume the same set of notations and assumptions in Theorem~\ref{thm_main_noiseless}.
  Suppose $\mat \Psi$ is a JL transform with parameter $\epsilon$.
Let $B:=\min_{\ell=1,...,k}\{\rho,r^{-1/2},\rho_\ell - \mu_\ell\}$. %, and $C$ be the constant in Theorem~\ref{thm_main_noiseless}~and~\ref{thm_main_noisy_rand}.
If the privacy parameter $\varepsilon$ satisfies
\begin{equation}
\varepsilon = \Omega\left(\sqrt{\frac{\mu_0\log(1/\delta)}{d}}\cdot\max\left\{\frac{\log^{5/4}{N}}{B^2}, \frac{\log^{3/2}{N}}{B^{11/2}p^{1/4}}\right\}\right),
\label{eq:varepsilon}
\end{equation}
then the solution to Lasso-SSC using obeys the subspace detection property with probability $1-8/N-\delta$.
\end{cor}

Corollary \ref{cor:utility} should be interpreted with care: though we place a lower bound condition on the privacy parameter $\varepsilon$, 
which might sound strange because $\varepsilon$ is usually specified by users,
such lower bound condition is only required to establish utility guarantees.
The algorithm itself is \emph{always} ($\varepsilon,\delta)$-differential private regardless of $\varepsilon$ values, as stated in Theorem \ref{thm:privacy}.
On the other hand, if the original data $\mat X$ is sufficiently incoherent (i.e., $\mu_0=O(1)$), 
the right-hand side of Eq.~(\ref{eq:varepsilon}) scales as $O(\mathrm{poly}(r)/\sqrt{d})$, which quickly approaches zero as the ambient dimension $d$ increases.
As a result, the lower bound in Eq.~(\ref{eq:varepsilon}) is a very mild condition on incoherent data matrix $\mat X$.

 The proof idea of Corollary \ref{cor:utility} is simple. We are now injecting artificial Gaussian noise to a compressed subspace clustering problem with fixed input, and Theorem~\ref{thm_main_noisy_rand} directly addresses that. All we have to do is to replace the geometric quantities in $\mu_\ell$ and $\rho_\ell$ by their respective bound after compression in Corollary~\ref{cor_subspace_incoherence_noiseless} and Lemma~\ref{lem_inradius_preservation}.
 The complete proof is deferred to Sec. \ref{subsec:proof-utility}.

 Before proceeding to the impossibility results we make some remarks on the condition of the privacy parameter $\varepsilon$.
 It can be seen that, smaller privacy parameter (i.e., higher degree of privacy protection) is possible on data sets with larger ``geometric gap''
 that makes subspace clustering relatively easy to succeed on such data sets.
 In addition, incoherence helps with privacy preservation, as $\varepsilon$ can be set at a smaller level on more incoherent data.

\subsection{Impossibility results}\label{sec:impossibility}

As we described in the main results, attribute-level differential privacy is a much weaker notion of privacy. 
A stronger privacy notion is \emph{user-level} differential privacy, where two neighboring databases $\mat X$ and $\mat X'$
differ by a \emph{column} rather than an entry, and hence the privacy of the entire attribute vector of each user in $\mat X$ is protected.
However, we show in this section that user-level differential privacy cannot be achieved if utility is measured in terms of (perfect) subspace detection property.
%While it is easy to handle a small group of attributes (in the order of $O(\sqrt{d/p})$ if we consider $B=O(1/r)$) by the composition rule, it does not protect any individual user's complete information.
%However, this is arguably the best we can do if our measure of utility is in terms of (perfect) subspace detection property.

We first give a formal definition of user-level differential privacy:
\begin{defn}[User-Level Differential Privacy]\label{defn:user-privacy}
We say a randomized algorithm $\cA: \R^{d\times N} \rightarrow \cO $ is $(\varepsilon,\delta)$-differential private at user level if
$$\Pr(\cA(\mat X)\in \cS) \leq e^{\varepsilon}\Pr(\cA(\mat X')\in \cS) + \delta$$
for any measurable outcome $\cS\subset \cO$, any $\mat X, \mat X' \in \cX^n$ that differs in only one column.
\end{defn}
Compared with the attribute differential privacy defined in \ref{defn:attribute-privacy}, the only difference 
is how $\mat X$ and $\mat X'$ may differ. Note that we can arbitrarily replace any single point in $\mat X$ with any $x\in \cX$, to form $\mat X'$.

The following proposition shows that user-level differential privacy cannot be preserved if perfect subspace detection or clustering is desired.
Its proof is placed in Sec.~\ref{subsec:proof-impossible}.
\begin{prop}\label{prop:impossible}
User-level differential privacy is NOT possible for any $0\leq \varepsilon<\infty$ using any privacy mechanism
if perfect subspace detection property or perfect clustering results are desired.
In addition,
If an algorithm achieves perfect clustering or subspace detection with probability $1-\delta$ for some $\delta\in(0,1)$, 
user-level differential privacy is NOT possible for any $\varepsilon < \log\left(\frac{1-\delta}{\delta}\right)$.
\end{prop}

Intuitively, the reason why attribute-level privacy is not subject to the impossibility result in Proposition \ref{prop:impossible} is because the privacy promise is much weaker:
even if perfect subspace detection or clustering results are presented, it is still possible to hide the information of a \emph{specific} attribute
of a user provided that the attribute values are distributed in a incoherent and near-uniform way.
On the other hand, change in a user's complete attribute file may often alter the cluster that user belongs to and eventually break perfect subspace clustering.
%Also our assumption that the columns are non-spiky ensures that perturbing any attribute of any user will not inject too much error. Intuitively, random projection and the injected dense Gaussian noise makes sure that it is not possible to identify any small changes in one attribute of a single user.

%To be fair, the same problem still exists, namely, differential privacy breaks whenever the clustering can be shown to be always correct.
%in that the user may still be identified if a user is completely replaced.
%What attribute-differential privacy ensures is that it is not possible to tell if a specific attribute of this user used in coming up with the result is actually the same or close to what it truly is.

User-level privacy for sparse subspace clustering and for privacy in general remains an important open problem. 
Some progress has been made in \citet{private-ssc} to address user-level private subspace clustering by weakening the utility guarantee from correct clustering
to approximately identifying underlying subspaces.
Nevertheless, the analysis in \citet{private-ssc} mostly focus on simpler algorithms like thresholding-based subspace clustering \citep{tsc} and cannot be easily generalized to 
state-of-the-art subspace clustering methods such as SSC \citep{ssc} or LRR \citep{lrr}.
%What we know for sure is that, we need to come up with a different/soft measure of utility other than exact clustering or subspace detection property.

\section{Proofs}

%\todo{YX: I did not carefully go through this section. Caught a few typos and grammar errors.  The equations should not have any issues, but I would suggest give the text in between equations another pass.}

Success condition for exact SSC was proved in \cite{sc-geometric} and was generalized to the noisy case in \cite{noisy-ssc}.
Below we cite Theorem 6 and Theorem 8 in \cite{noisy-ssc} for a success condition of Lasso SSC.
In general, Lasso SSC succeeds when there is a sufficiently large gap between subspace incoherence and inradius.
Results are restated below, with minor simplification in our notation.
\begin{thm}[{\citealp[Theorem 6 and 8]{noisy-ssc}}]
Suppose $\mat X=\mat Y+\mat Z$ where $\mat Y$ is the uncorrupted data matrix and $\mat Z=(\vct z_1,\cdots,\vct z_N)$ is a deterministic noise matrix
that satisfies $\max_{i=1}^N{\|\vct z_i\|_2} \leq \eta$.
Define $\rho := \min_\ell{\rho_\ell}$.
%If $\mu_\ell < \rho_\ell$ for $\ell=1,2,\cdots,k$ and
If
\begin{equation}
%\eta \leq \min_{\ell=1,\cdots,k}{\frac{\rho(\rho_\ell-\mu_\ell)}{2\rho_\ell^2+6\rho_\ell+2}},
\eta \leq \min_{\ell=1,\cdots,k}{\frac{\rho(\rho_\ell-\mu_\ell)}{7\rho_\ell+2}},
\label{eq_snr_ratio}
\end{equation}
then subspace detection property holds for the Lasso SSC algorithm in Eq. (\ref{eq_lasso_ssc}) if the regularization coefficient $\lambda$
is in the range
\begin{equation}
\max_{\ell=1,\cdots,k}{\frac{\eta(1+\eta)(2+\rho_\ell)}{\rho_\ell-\mu_\ell-2\eta}}
< \lambda
< \rho - 2\eta - \eta^2.
\label{eq_noisy_lambda_range}
\end{equation}
In addition, if $\mat Z_{ij}\sim\nml(0, \sigma_{ij}^2/d)$ are independent Gaussian noise with variance $\sigma^2:=\max_{i,j}\sigma_{ij}^2$ satisfying
\begin{equation}
\sqrt{\frac{\log N}{d}}\sigma(1+\sigma) < C\min_{\ell=1,\cdots,k}\left\{\rho,r^{-1/2},\rho_\ell-\mu_\ell\right\}
\label{eq_sigma_bound}
\end{equation}
for sufficiently small constant $C\geq 1/80$, then with probability at least $1-\frac{10}{N}$ the subspace detection property holds
%for the Lasso SSC algorithm in Eq. (\ref{eq_lasso_ssc})
if $\lambda$ is in the range
\begin{equation}
\frac{C_1\sigma(1+\sigma)}{\rho_\ell-\mu_\ell}\sqrt{\frac{\log N}{d}} < \lambda <  \rho- C_2\sigma(1 +\sigma)\sqrt{\frac{\log N}{d}}.
\label{eq_sigma_lambda_bound}
\end{equation}
Here $C_1\leq 80$ and $C_2\leq 20$ are absolute constants.

%\red{\textbf{[Not so useful in the statement?]}In addition, for the noiseless setting ($\eta = 0$) the Lasso SSC algorithm succeeds if $\mu_\ell < \rho_\ell$ for all subspace $\mathcal U^{(\ell)}$
%and $\lambda < \rho$.}
\label{thm_sep_noisy}
\end{thm}

\subsection{Proof of Theorem \ref{thm_main_noiseless}}

We first bound the perturbation of dual directions when the data are noiseless.
\begin{lem}[Perturbation of dual directions, the noiseless case]
Assume $0<\lambda<1$.
Fix a column $\vct x$ in $\mat X$ with dual direction $\vct\nu=\vct\nu(\vct x)$ and $\vct v=\vct v(\vct x)$ defined in Eq. (\ref{eq_lasso_dual}) and (\ref{eq_subspace_incoherence}).
Let $\widetilde{\mat X}$ denote the projected data matrix $\mat\Psi\mat X$
and $\widetilde{\mat X}'$ denote the normalized version of $\widetilde{\mat X}$.
Suppose $\vct\nu^*$ and $\vct v^*$ are computed using the normalized projected data matrix $\widetilde{\mat X}'$.
If $\mat\Psi$ satisfies Eq. (\ref{eq_well_behave}, \ref{eq_well_behave_jl}) with parameter $\epsilon$
and $\epsilon < 1/2\max(1,\|\vct\nu\|^2)$ then the following holds for all $\vct w\in\mat X\backslash\mat X^{(\ell)}$:
\begin{equation}
\big| \langle\vct v,\vct w\rangle - \langle\vct v^*,\tilde{\vct w}'\rangle\big| = O\left(\epsilon^{1/2}\lambda^{-3/4}\right).
%\leq 64\sqrt{\epsilon/\lambda} + 2\epsilon.
\label{eq_perturbation_nu}
\end{equation}
\label{lem_subspace_incoherence_noiseless}
\end{lem}

\begin{proof}
Fix $\ell\in[k]$ and one column $\vct x_i$ in $\mat X$.
Let $\mathcal U^{(\ell)}$ and $\widetilde{\mathcal U}^{(\ell)}$ denote the low-rank subspaces to which $\vct x_i$ belongs 
before and after compression.
That is, $\widetilde{\mathcal U}^{(\ell)} = \{\mat\Psi\vct x: \vct x\in\mathcal U^{(\ell)}\}$.

First note that $(1-\lambda/2)^2\leq \|\vct\nu\|^2 \leq 2/\lambda$.
$\|\vct\nu\|\geq 1-\lambda/2$ because $\langle\vct x,\vct\nu\rangle - \lambda/2\cdot \|\vct\nu\|^2 \leq \langle\vct x,\vct\nu\rangle\leq \|\vct\nu\|$
and putting $\vct\nu = \vct x$ we obtain a solution with value $1-\lambda/2$.
On the other hand, $\langle\vct x,\vct\nu\rangle - \lambda/2\cdot \|\vct\nu\|^2 \leq \|\vct\nu\| - \lambda/2\cdot \|\vct\nu\|^2$
and putting $\vct\nu = \vct 0$ we obtain a solution with value 0.
Also, under the noiseless setting $\vct\nu\in\mathcal U^{(\ell)}$, if $\vct x\in\mathcal U^{(\ell)}$.

Define 
$$\tilde{\vct\nu}' = \frac{\sqrt{1-\epsilon}}{1+\epsilon\max(1,\|\vct\nu\|^2)}\cdot\tilde{\vct\nu},$$
where $\tilde{\vct\nu} = \mat\Psi\vct\nu$.
Let $f(\vct\nu) = \langle\vct\nu,\vct x\rangle - \frac{\lambda}{2}\|\vct\nu\|_2^2$
and $\tilde f(\tilde{\vct\nu}') = \langle\tilde{\vct\nu}', \tilde{\vct x}'\rangle - \frac{\lambda}{2}\|\tilde{\vct\nu}'\|_2^2$
denote the values of the optimization problems.
The first step is to prove that $\tilde{\vct\nu}$ is feasible and nearly optimal to the projected optimization problem;
that is, $\tilde f(\tilde{\vct\nu}')$ is close to $\tilde f(\vct\nu^*)$.

We first show that $\tilde{\vct\nu}'$ is a feasible solution with high probability.
By Proposition \ref{prop_well_behave}, the following bound on $|\tilde{\vct x}_i^\top\tilde{\vct\nu}|$ holds:
\begin{equation}
|\tilde{\vct x}_i^\top\tilde{\vct\nu}| \leq |\vct x_i^\top\vct\nu| + \epsilon\cdot\frac{\|\vct x_i\|^2+\|\vct\nu\|^2}{2} \leq 1 +\epsilon\max(1,\|\vct\nu\|^2).
\quad\forall \vct x_i\in\mat X.
\end{equation}
Furthermore, with probability $\geq 1-\delta$
\begin{equation}
\|\tilde{\vct x}_i\|_2^2 \geq (1-\epsilon)\|\vct x_i\|_2^2 = 1-\epsilon.
\end{equation}
Consequently, by the definition of $\tilde{\vct\nu}'$ one has
\begin{equation}
\|\widetilde{\mat X}'^\top\tilde{\vct\nu}'\|_{\infty}
\leq \frac{1}{\sqrt{1-\epsilon}}\cdot \frac{\sqrt{1-\epsilon}}{1+\epsilon\max(1,\|\vct\nu\|^2)}\|\widetilde{\mat X}^\top\tilde{\vct\nu}\|_{\infty}
\leq 1.
\end{equation}

Next, we compute a lower bound on $\tilde f(\tilde{\vct\nu}')$, which should serve as a lower bound for $\tilde f(\vct\nu^*)$
because $\vct\nu^*$ is the optimal solution to the dual optimization problem on the projected data.
We first remark that $\langle\vct x,\vct\nu\rangle\geq 1-\lambda/2$ due to the optimality condition of the dual problem at $\vct\nu=\vct x$ and hence
$$
\langle\tilde{\vct x},\tilde{\vct\nu}\rangle \geq \langle\vct x,\vct\nu\rangle - \epsilon\cdot \frac{\|\vct x\|^2+\|\vct\nu\|^2}{2} \geq 1-\frac{\lambda}{2}-\epsilon\max(1,\|\vct\nu\|^2) \geq 0,
$$
where the last inequality is due to the assumption that $\lambda\in(0,1)$ and $\epsilon<1/2\max(1,\|\vct\nu^2\|)$.
Consequently, we have the following chain of inequalities:
\begin{eqnarray}
\tilde f(\tilde{\vct\nu}')
&=& \langle\tilde{\vct x}', \tilde{\vct\nu}'\rangle - \frac{\lambda}{2}\|\tilde{\vct\nu}'\|_2^2\nonumber\\
&\geq& \sqrt{\frac{1-\epsilon}{1+\epsilon}}\frac{\langle\tilde{\vct x},\tilde{\vct\nu}\rangle}{1+\epsilon\max(1,\|\vct\nu\|^2)} - \frac{\lambda}{2}(1-\epsilon)\|\tilde{\vct\nu}\|_2^2\nonumber\\
&\geq& (1-\epsilon)(1-\epsilon\max(1,\|\vct\nu\|^2))\left(\langle\vct x,\vct\nu\rangle - \epsilon\max(1,\|\vct\nu\|^2)\right) - \frac{\lambda}{2}(1-\epsilon)(1+\epsilon)\|\vct\nu\|^2\nonumber\\
&\geq& (1-\epsilon)(\langle\vct x,\vct\nu\rangle - \epsilon\max(1,\|\vct\nu\|^2)) - \epsilon\max(1,\|\vct\nu\|^2)\cdot\langle\vct x,\vct\nu\rangle - \frac{\lambda}{2}\|\vct\nu\|^2\nonumber\\
&\geq& \langle\vct x,\vct\nu\rangle - \epsilon\max(1,\|\vct\nu\|^2) - \epsilon\langle\vct x,\vct\nu\rangle - \epsilon\max(1,\|\vct\nu\|^2)\cdot\langle\vct x,\vct\nu\rangle - \frac{\lambda}{2}\|\vct\nu\|^2\nonumber\\
&\geq& \langle\vct x,\vct\nu\rangle - \epsilon\max(1,\|\vct\nu\|^2) - \epsilon\|\vct\nu\| - \epsilon\max(1,\|\vct\nu\|^2)\cdot \|\vct\nu\| - \frac{\lambda}{2}\|\vct\nu\|^2\nonumber\\
&\geq& f(\vct\nu) - 3\epsilon\max(1,\|\vct\nu\|,\|\vct\nu\|^2, \|\vct\nu\|^3)\\
&\geq& f(\vct\nu) - 3\epsilon\max(1,\|\vct\nu\|^3).
\end{eqnarray}
%Hence, $\tilde f(\vct\nu^*) \geq \tilde f(\tilde{\vct\nu}') \geq f(\vct\nu)-2\epsilon\max(1,\|\vct\nu\|)$.
On the other hand, since $\vct\nu^*\in\widetilde{\mathcal U}^{(\ell)}$, there exists $\bar{\vct\nu}\in\mathcal U^{(\ell)}$ such that $\vct\nu^* = \mat\Psi\bar{\vct\nu}$.
Let $\bar{\vct\nu}'$ be a scaled version of $\bar{\vct\nu}$ so that it is a feasible solution to the optimization problem in Eq. (\ref{eq_lasso_dual}) before projection.
Using essentially similar analysis one can show that $f(\bar{\vct\nu}') \geq \tilde f(\vct\nu^*) - 3\epsilon\max(1,\|\vct\nu^*\|^3)$.
Consequently, the following bound on the gap between $\tilde f(\tilde{\vct\nu}')$ and $\tilde f(\vct\nu^*)$ holds:
\begin{equation}
\big|\tilde f(\tilde{\vct\nu}') - \tilde f(\vct\nu^*)\big| \leq 6\epsilon\max(1,\|\vct\nu\|^3,\|\vct\nu^*\|^3).
\end{equation}

Because the dual problem in Eq. (\ref{eq_lasso_dual}) is strongly convex with parameter $\lambda$ (this holds for both the projected and the original problem),
we can bound the perturbation of dual directions $\|\tilde{\vct\nu}'-\vct\nu^*\|$ by the bounds on their values $|\tilde f(\tilde{\vct\nu}')-\tilde f(\vct\nu^*)|$ as
\begin{equation}
\|\tilde{\vct\nu}' - \vct\nu^*\|_2 \leq \sqrt{\frac{2|\tilde f(\tilde{\vct\nu}')-\tilde f(\vct\nu^*)|}{\lambda}} \leq \sqrt{\frac{12\epsilon\max(1,\|\vct\nu\|^3,\|\vct\nu^*\|^3)}{\lambda}}.
\end{equation}

Next, note that $\tilde{\vct\nu}',\vct\nu^*\in\widetilde{\mathcal U}^{(\ell)}$.
Also note that for any two vector $\vct a,\vct b$ the following holds:
\begin{eqnarray*}
\left\|\frac{\vct a}{\|\vct a\|} - \frac{\vct b}{\|\vct b\|}\right\|
&=& \left\|\frac{\vct a}{\|\vct a\|} - \frac{\vct b}{\|\vct a\|} + \frac{\vct b}{\|\vct a\|} - \frac{\vct b}{\|\vct b\|}\right\|\\
&\leq& \frac{\|\vct a-\vct b\|}{\|\vct a\|} + \frac{\|\vct b\|\cdot \big|\|\vct a\|-\|\vct b\|\big|}{\|\vct a\|\|\vct b\|}\\
&\leq& \frac{\|\vct a-\vct b\|}{\|\vct a\|} + \frac{\|\vct a-\vct b\|}{\|\vct a\|}\\
&=& \frac{2\|\vct a-\vct b\|}{\|\vct a\|}.
\end{eqnarray*}
By symmetry we also have $\|\frac{\vct a}{\|\vct a\|} - \frac{\vct b}{\|\vct b\|}\| \leq \frac{2\|\vct a-\vct b\|}{\|\vct b\|}$.
Therefore,
\begin{equation}
\left\|\frac{\vct a}{\|\vct a\|} - \frac{\vct b}{\|\vct b\|}\right\| \leq \frac{2\|\vct a-\vct b\|}{\max(\|\vct a\|,\|\vct b\|)}.
\end{equation}

Now we can bound $\|\tilde{\vct v}'-\vct v^*\|$ as follows:
\begin{eqnarray*}
\|\tilde{\vct v}'-\vct v^*\|
&=& \left\|\frac{\tilde{\vct\nu}'}{\|\tilde{\vct\nu}'\|} - \frac{\vct\nu^*}{\|\vct\nu^*\|}\right\|\\
&\leq& \frac{2\|\tilde{\vct\nu}'-\vct\nu^*\|}{\max(\|\tilde{\vct\nu}'\|,\|\vct\nu^*\|)}
\leq \frac{2\|\tilde{\vct\nu}'-\vct\nu^*\|}{\max(\|\vct\nu\|/4, \|\vct\nu^*\|)}\\
&\leq& \frac{16\sqrt{3\epsilon\max(1,\|\vct\nu\|^3, \|\vct\nu^*\|^3)}}{\sqrt{\lambda}\max(\|\vct\nu\|,\|\vct\nu^*\|)}
\leq 16\sqrt{\frac{3\epsilon}{\lambda}\max\left(\|\vct\nu\|, \|\vct\nu^*\|,\frac{1}{\|\vct\nu\|^2},\frac{1}{\|\vct\nu^*\|^2}\right)}\\
&\leq& 16\sqrt{\frac{3\epsilon}{\lambda}\left(\frac{1}{(1-\lambda/2)^2} + \sqrt{\frac{2}{\lambda}}\right)}\\
&\leq& 16\sqrt{\frac{3\epsilon}{\lambda}\left(\sqrt{\frac{16}{\lambda}} + \sqrt{\frac{2}{\lambda}}\right)} = O\left(\epsilon^{1/2}\lambda^{-3/4}\right).
\end{eqnarray*}
where the last line comes from the fact that $\lambda\in(0,1)$.

Note that after normalization $\tilde{\vct v}'$ is exactly the same with $\tilde{\vct v}$.
Subsequently, for any $\vct y\in\mat X\backslash\mat X^{(\ell)}$ we have
\begin{eqnarray*}
\big|\langle\vct v,\vct y\rangle - \langle\vct v^*,\tilde{\vct y}'\rangle\big|
&\leq& \big|\langle\tilde{\vct v}',\tilde{\vct y}'\rangle - \langle\vct v^*,\tilde{\vct y}'\rangle\big| + \big|\langle\tilde{\vct v}',\tilde{\vct y}'\rangle - \langle\vct v,\vct y\rangle\big|\\
&\leq& \|\tilde{\vct v}'-\vct v^*\|\|\tilde{\vct y}'\| + \big|\langle\tilde{\vct v},\tilde{\vct y}'\rangle - \langle\vct v,\vct y\rangle\big|\\
&\leq& \|\tilde{\vct v}'-\vct v^*\| + \bigg|\frac{1}{\|\mat\Psi\vct v\|\|\mat\Psi\vct y\|}\langle\mat\Psi\vct v,\mat\Psi\vct y\rangle - \langle\vct v,\vct y\rangle\bigg|\\
&\leq& O\left(\epsilon^{1/2}\lambda^{-3/4}\right) + \left(1-\frac{1}{\|\mat\Psi\vct v\|\|\mat\Psi\vct y\|}\right)\|\mat\Psi\vct v\|\|\mat\Psi\vct y\| + \big|\langle\mat\Psi\vct v,\mat\Psi\vct y\rangle-\langle\vct v,\vct y\rangle\big|\\
&\leq& O\left(\epsilon^{1/2}\lambda^{-3/4}\right) + \left(1-\frac{1}{1+\epsilon}\right)(1+\epsilon) + \epsilon\\
&=& O\left(\epsilon + \epsilon^{1/2}\lambda^{-3/4}\right) =O\left(\epsilon^{1/2}\lambda^{-3/4}\right).
\end{eqnarray*}

\end{proof}

As a simple corollary, perturbation of subspace incoherence can then be bounded as in Corollary \ref{cor_subspace_incoherence_noiseless}.
\begin{cor}[Perturbation of subsapce incoherence, the noiseless case]
Assume the same notations in Lemma \ref{lem_subspace_incoherence_noiseless}.
\label{cor_subspace_incoherence_noiseless}
Let $\mu_\ell$ and $\tilde\mu_\ell$ be the subspace incoherence of the $\ell$th subspace before and after dimension reduction.
Then the following holds for all $\ell=1,\cdots,k$:
\begin{equation}
\tilde\mu_\ell = \mu_\ell + O\left(\epsilon^{1/2}\lambda^{-3/4}\right).
\label{eq_perturbation_mu}
\end{equation}
\end{cor}

The following lemma bounds the perturbation of inradius for each subspace.
\begin{lem}[Perturbation of inradius]
\label{lem_inradius_preservation}
Fix $\ell\in\{1,\cdots,k\}$ and $\delta,\epsilon > 0$.
Let $\mat X = \mat X^{(\ell)} = (\vct x_1,\cdots,\vct x_{N_\ell})\subseteq \mathcal U^{(\ell)}$
be the noiseless $d\times N_\ell$ matrix with all columns belonging to $\mathcal U^{(\ell)}$ with unit two norm.
%Denote $\mathcal P(\mat Y) = \conv\{\pm\vct y_1,\cdots,\pm\vct y_m\}$ to be the symmetric convex polytope formed by all columns in $\mat Y$
%and $r(\mathcal P(\mat Y))$ its inradius.
Suppose $\widetilde{\mat X} = \mat\Psi\mat X\in\mathbb R^{p\times N_{\ell}}$ is the projected matrix
and $\widetilde{\mat X'}$ scales every column in $\widetilde{\mat X}$ so that they have unit norm.
Let $\rho_\ell$ and $\tilde{\rho_\ell}$ be the inradius of subspace $\mathcal U^{(\ell)}$ before and after dimensionality reduction,
defined on $\mat X$ and $\widetilde{\mat X}'$ respectively.
If $\mat\Psi$ satisfies Eq. (\ref{eq_well_behave},\ref{eq_well_behave_jl}) with parameter $\epsilon$
then with probability $\geq 1-\delta$ the following holds:
\begin{equation}
\tilde\rho_\ell \geq \rho_\ell/(1+\epsilon).
\end{equation}
\end{lem}

\begin{proof}
For notational simplicity re-define $\mat X=\mat X_{(-i)}$ and $\widetilde{\mat X}'=\widetilde{\mat X}_{(-i)}'$ for some fixed data point $\vct x_i^{(\ell)}$.
Let $\mathcal C$, $\widetilde{\mathcal C}$ be the largest Euclidean balls inscribed in $\mathcal Q(\mat X)$ and $\mathcal Q(\widetilde{\mat X})$.
Since both $\mathcal Q(\mat X)$ and $\mathcal Q(\widetilde{\mat X})$ are symmetric convex bodies with respect to the origin,
the centers of $\mathcal C$ and $\widetilde{\mathcal C}$ are the origin.
Let $\tilde{\vct c}$ be any point in $\widetilde{\mathcal C}\cap\partial\mathcal Q(\widetilde{\mat X})$.
By definition,  $r(\mathcal Q(\widetilde{\mat X})) = \|\tilde{\vct c}\|$.
Since $\tilde{\vct c}\in\widetilde{\mathcal U}^{(\ell)}$, we can find $\vct c\in\mathcal U^{(\ell)}$ such that $\tilde{\vct c} = \mat\Psi\vct c$.
By Proposition \ref{prop_well_behave}, we have (with probability $\geq 1-\delta$)
\begin{equation}
\|\tilde{\vct c}\| \geq \frac{1}{\sqrt{1+\epsilon}}\|\vct c\|.
\end{equation}
On the other hand, $\vct c$ is not contained in the interior of $\mathcal Q(\mat X)$.
Otherwise, we can find a scalar $a > 1$ such that $a\vct c\in\mathcal Q(\mat X)$ and hence $a\tilde{\vct c}\in\mathcal Q(\widetilde{\mat X})$,
contradicting the fact that $\tilde{\vct c}\in\partial\mathcal Q(\widetilde{\mat X})$.
Consequently, we have $\|\vct c\| \geq r(\mathcal Q(\mat X))$ by definition.
Therefore,
\begin{equation}
r(\mathcal Q(\widetilde{\mat X})) = \|\tilde{\vct c}\| \geq \frac{1}{\sqrt{1+\epsilon}}\|\vct c\| \geq \frac{r(\mathcal Q(\mat X))}{\sqrt{1+\epsilon}}.
\end{equation}

Next, we need to lower bound $r(\mathcal Q(\widetilde{\mat X}'))$ in terms of $r(\mathcal Q(\widetilde{\mat X}))$.
This can be easily done by noting that the maximum column norm in $\widetilde{\mat Y}$ is upper bounded by $\sqrt{1+\epsilon}$.
Consequently, we have
\begin{equation}
r(\mathcal Q(\widetilde{\mat X}')) \geq r\left(\mathcal Q\left(\frac{1}{\sqrt{1+\epsilon}}\widetilde{\mat X}\right)\right)\geq \frac{r(\mathcal Q(\mat X))}{1+\epsilon}.
\end{equation}

\end{proof}

With the perturbation of both subspace incoherence and inradius under dimensionality reduction,
we are now able to prove Theorem \ref{thm_main_noiseless}.
\begin{proof}[Proof of Theorem \ref{thm_main_noiseless}]
Let $\tilde\mu_{\ell},\tilde{\rho_{\ell}}$ denote the subspace incoherence and inradius of subspace $\mathcal U^{(\ell)}$ after dimensionality reduction.
Theorem \ref{thm_sep_noisy} shows that Lasso SSC satisfies the subspace detection property if $\tilde\mu_\ell < \tilde\rho_\ell$ for every $\ell$
and $\lambda < \tilde\rho$.
By Lemma \ref{lem_inradius_preservation}, $\tilde\rho \geq \rho/2$ with high probability
and hence $0<\lambda<\rho/2$ is sufficient to guarantee that $\lambda<\tilde\rho$ with high probability.
Note also that $\tilde\rho_\ell \geq \frac{\rho_{\ell}}{1+\epsilon} \geq \rho_{\ell}(1-\epsilon)$.
Subsequently, the following inequality yields $\tilde\mu_\ell<\tilde\rho_\ell$ for every $\ell$:
\begin{equation}
\mu_\ell + O\left(\epsilon^{1/2}\lambda^{-3/4}\right) + \rho_\ell\epsilon < \rho_\ell,\quad\forall \ell=1,\cdots,k.
\label{eq_pr1_main}
\end{equation}
Note that when $\epsilon$ is sufficiently small, the $\rho_\ell\epsilon$ term is overwhelmingly smaller than $O\left(\epsilon^{1/2}\lambda^{-3/4}\right)$.
Therefore, taking $\epsilon$ as
$$
\epsilon = O\left(\Delta^2\cdot \lambda^{3/2}\right)
$$
satisfies Eq. (\ref{eq_pr1_main}), where $\Delta = \min_\ell(\rho_\ell-\mu_\ell)$ is the critical geometric gap as defined in the main text.
Finally, we note that $\epsilon < \lambda/4$ is sufficient to guarantee the $\epsilon < 1/2\|\vct\nu\|^2$ condition in Lemma \ref{lem_subspace_incoherence_noiseless}
because $\|\vct\nu\|^2\leq 2/\lambda$.
\end{proof}

\subsection{Proof of Theorem \ref{thm_main_noisy}}

When the input matrix is corrupted with deterministic (adversarial) noise, Lemma \ref{lem_inradius_preservation} remains unchanged
because the inradius is defined in terms of the noiseless data matrix $\mat Y$.
In fact we could simply replace every occurrence of $\mat X$ with $\mat Y$ in the proof of Lemma \ref{lem_inradius_preservation}
to obtain the same perturbation bound for inradius defined for noisy SSC, as in Definition \ref{defn_inradius_noisy}.
Therefore, we only need to prove a noisy version of Lemma \ref{lem_subspace_incoherence_noiseless}
that bounds the perturbation of dual directions.

\begin{lem}[Perturbation of dual directions under deterministic noise]
Assume $0<\lambda<1$.
Suppose $\mat X=\mat Y+\mat Z$ where $\mat Y$ is the uncorrupted data matrix and $\mat Z$ is the noise matrix
with $\max_{i=1,\cdots,n}{\|\vct z_i\|_2} \leq \eta$.
Fix a column $\vct x$ with dual direction $\vct\nu$ and $\vct v$ defined in Eq. (\ref{eq_lasso_dual}) and (\ref{eq_subspace_incoherence}).
Suppose $\widetilde{\mat Y} = \mat\Psi\mat Y$ is the projected noiseless data matrix and $\widetilde{\mat Y}'$ is the normalized version of $\widetilde{\mat Y}$.
Let $\widetilde{\mat X}' = \widetilde{\mat Y}'+\widetilde{\mat Z}'$ be the noisy observation after projection,
where $\widetilde{\mat Z}'$ is the normalized version of the projected noise matrix $\widetilde{\mat Z}=\mat\Psi\mat Z$.
If $\mat\Psi$ satisfies Eq. (\ref{eq_well_behave},\ref{eq_well_behave_jl}) with parameter $\epsilon$
and $(\epsilon+4\eta) < 1/2\max(1,\|\vct\nu\|^2)$ then the following holds for all $\vct w\in\mat X\backslash\mat X^{(\ell)}$:
\begin{equation}
\big| \langle\vct v,\vct w\rangle - \langle\vct v^*,\tilde{\vct w}'\rangle\big| = O\left(\eta\rho_\ell^{-1/2} + (\epsilon+\eta)^{1/2}\lambda^{-3/4}\right).
\label{eq_perturbation_nu_noisy}
\end{equation}
\label{lem_subspace_incoherence_noisy}
\end{lem}

\begin{proof}

Fix $\ell\in\{1,2,\cdots,k\}$ and a particular column $\vct x=\vct x_i$.
Suppose $\vct\nu$ is the optimal solution to the original dual problem in Eq. (\ref{eq_lasso_dual}).
Define $\vct\nu_{\parallel} = \mathcal P_{\mathcal U^{(\ell)}}\vct\nu$ and $\vct\nu_\perp = \mathcal P_{\mathcal U^{(\ell)^\perp}}\vct\nu$.
Let $f(\cdot)$ be the objective value of the dual problem under a specific solution.
Then it is easy to observe that
\begin{eqnarray}
f(\vct\nu_\parallel)
&=& \langle\vct\nu_\parallel, \vct x\rangle - \frac{\lambda}{2}\|\vct\nu_\parallel\|_2^2\nonumber\\
&\geq& \langle\vct\nu_\parallel,\vct x\rangle - \frac{\lambda}{2}\|\vct\nu\|_2^2\nonumber\\
&=& \langle\vct\nu,\vct x\rangle - \langle\vct\nu_\perp,\vct x\rangle - \frac{\lambda}{2}\|\vct\nu\|_2^2\nonumber\\
&=& \langle\vct\nu,\vct x\rangle - \langle\vct\nu_\perp,\vct x_\perp\rangle - \frac{\lambda}{2}\|\vct\nu\|_2^2\nonumber\\
 &=& f(\vct\nu) - \langle \vct x_{\perp}, \vct\nu_{\perp}\rangle \nonumber\\
 &\geq& f(\vct\nu) - \eta\|\vct\nu_\perp\|_2.\nonumber
\end{eqnarray}
We then cite the following upper bound for $\|\vct\nu_\perp\|$, which appears as Eq. (5.16) in \citep{noisy-ssc}.
\begin{equation}
\|\vct\nu_\perp\|_2 \leq \lambda\eta\left(\frac{1}{r(\mathcal Q(\mat Y_{-i}^{(\ell)}))} + 1\right) \leq \frac{2\lambda\eta}{\rho_\ell}.
\label{upper_bound_nu2}
\end{equation}

Let $\tilde{\vct\nu} = \mat\Psi\vct\nu_\parallel$ and 
$$\tilde{\vct\nu}' = \frac{\sqrt{1-\epsilon}}{1+(\epsilon+4\eta)\max(1,\|\vct\nu\|^2)}\cdot\tilde{\vct\nu}.$$
It is easy to verify that $\tilde{\vct\nu}'$ is a feasible solution to the projected dual problem.
To see this, note that
\begin{eqnarray*}
\langle\widetilde{\vct x}, \widetilde{\vct\nu}\rangle
&=& \langle\mat\Psi\vct x_\parallel, \mat\Psi\vct\nu_\parallel\rangle + \langle\mat\Psi\vct x_\perp,\mat\Psi\vct\nu_\parallel\rangle\\
&\leq& \langle\mat\Psi\vct x_\parallel, \mat\Psi\vct\nu_\parallel\rangle + (1+\epsilon)\|\vct x_\perp\|\|\vct\nu_\parallel\|\\
&\leq& \langle\mat\Psi\vct x_\parallel, \mat\Psi\vct\nu_\parallel\rangle + (1+\epsilon)\eta\|\vct\nu\|\\
&\leq& \langle\vct x_\parallel, \vct\nu_\parallel\rangle + \epsilon\cdot\frac{\|\vct x_\parallel\|^2 + \|\vct\nu_\parallel\|^2}{2} + (1+\epsilon)\eta\|\vct\nu\|^2\\
&\leq& \langle\vct x_\parallel, \vct\nu_\parallel\rangle + \epsilon\cdot\frac{\|\vct x\|^2 + \eta^2 + \|\vct\nu_\parallel\|^2}{2} + (1+\epsilon)\eta\|\vct\nu\|^2\\
&\leq& \langle\vct x_\parallel,\vct\nu_\parallel\rangle + \epsilon\max(1,\|\vct\nu\|^2) + \frac{\epsilon}{2}\eta + (1+\epsilon)\eta\|\vct\nu\|\\
&\leq& 1 + \eta\|\vct\nu_\perp\| + \epsilon\max(1,\|\vct\nu\|^2) + \epsilon\eta + (1+\epsilon)\eta\|\vct\nu\|\\
&\leq& 1 + \epsilon\eta + (2+\epsilon)\eta\|\vct\nu\| + \epsilon\max(1,\|\vct\nu\|^2)\\
&\leq& 1 + (\epsilon+4\eta)\max(1,\|\vct\nu\|,\|\vct\nu\|^2)\\
&\leq& 1 + (\epsilon+4\eta)\max(1,\|\vct\nu\|^2).
\end{eqnarray*}
In the above derivation, for the third and fourth lines we use the fact that $\mat\Psi$ is well behaved and $\vct x_\perp=\vct z$, $\vct\nu_\parallel\in\mathcal U^{(\ell)}$.
To see why the sixth line holds, notice that
$$
\langle\vct x,\vct\nu\rangle = \langle\vct x_\parallel,\vct\nu_\parallel\rangle + \langle\vct x_\perp,\vct\nu_\perp\rangle
$$
and hence
$$
\langle\vct x_\parallel,\vct\nu_\parallel\rangle \leq \langle\vct x,\vct\nu\rangle + \|\vct x_\perp\|\|\vct\nu_\perp\| \leq 1 + \eta\|\vct\nu_\perp\|.
$$
Consequently,
$$
\|\widetilde{\mat X}'^\top\widetilde{\vct\nu}'\|_{\infty}
\leq \frac{1}{\sqrt{1-\epsilon}}\cdot \frac{\sqrt{1-\epsilon}}{1+(\epsilon+4\eta)\max(1,\|\vct\nu\|^2)}\cdot \|\widetilde{\mat X}^\top\widetilde{\vct\nu}\|_{\infty} \leq 1.
$$

Next,
define $\eta' := \max_{i=1,\cdots,n}\|\tilde{\vct z}_i\|_2$.
Since $\mat\Psi$ is well behaved, $\eta' \leq \sqrt{1+\epsilon}\eta$ with high probability.
In addition, $\langle\tilde{\vct y},\tilde{\vct \nu}\rangle\geq 0$ with high probability because
\begin{eqnarray*}
\langle\tilde{\vct y},\tilde{\vct\nu}\rangle 
&\geq& \langle\vct y,\vct\nu\rangle - \epsilon\cdot \frac{\|\vct y\|^2 + \|\vct\nu\|^2}{2}\\
&=& \langle\vct x,\vct\nu\rangle - \epsilon\cdot\frac{\|\vct y\|^2 + \|\vct\nu\|^2}{2} - \langle\vct z,\vct\nu\rangle\\
&\geq& \langle\vct x,\vct\nu\rangle - \epsilon\max(1,\|\vct\nu\|^2) - \eta\|\vct\nu\|\\
&\geq& 0,
\end{eqnarray*}
where in the last inequality we use the fact that $\langle\vct x,\vct\nu\rangle\geq 1-\lambda/2$ and the assumption that $\lambda\in(0,1)$ and $(\epsilon+4\eta)\max(1,\|\vct\nu\|^2) < 1/2$.
Applying essentially the same chain of argument as in the proof of Lemma \ref{lem_subspace_incoherence_noiseless}
we obtain
\begin{eqnarray*}
\tilde f(\tilde{\vct\nu}')
&=& \langle\tilde{\vct x}', \tilde{\vct\nu}'\rangle - \frac{\lambda}{2}\|\tilde{\vct\nu}'\|_2^2\\
&=& \langle\tilde{\vct y}', \tilde{\vct\nu}'\rangle + \langle\tilde{\vct z}', \tilde{\vct\nu}'\rangle - \frac{\lambda}{2}\|\tilde{\vct\nu}'\|_2^2\\
&\geq& \sqrt{\frac{1-\epsilon}{1+\epsilon}}\frac{\langle\tilde{\vct y},\tilde{\vct\nu}\rangle}{1+(\epsilon+5\eta)\max(1,\|\vct\nu\|^2)} + \langle\tilde{\vct z}',\tilde{\vct\nu}'\rangle - \frac{\lambda}{2}\|\tilde{\vct\nu}'\|_2^2\\
&\geq& (1-\epsilon)(1-(\epsilon+5\eta)\max(1,\|\vct\nu\|^2))(\langle\vct y,\vct\nu_\parallel\rangle - (\epsilon+5\eta)\max(1,\|\vct\nu\|^2)) +\langle\tilde{\vct z}',\tilde{\vct\nu}'\rangle - \frac{\lambda}{2}\|\tilde{\vct\nu}'\|_2^2\\
&\geq& (1-\epsilon)(\langle\vct y,\vct\nu_\parallel\rangle - (\epsilon+5\eta)\max(1,\|\vct\nu\|^2)) - (\epsilon+5\eta)\max(1,\|\vct\nu\|^2)\cdot\langle\vct y,\vct\nu_\parallel\rangle +\langle\tilde{\vct z}',\tilde{\vct\nu}'\rangle - \frac{\lambda}{2}\|\tilde{\vct\nu}'\|_2^2\\
&\geq& \langle\vct y,\vct\nu_\parallel\rangle - \epsilon\|\vct\nu_\parallel\| - (\epsilon+5\eta)\max(1,\|\vct\nu\|^2) - (\epsilon+5\eta)\max(1,\|\vct\nu\|^2)\cdot\langle\vct y,\vct\nu_\parallel\rangle +\langle\tilde{\vct z}',\tilde{\vct\nu}'\rangle - \frac{\lambda}{2}\|\tilde{\vct\nu}'\|_2^2\\
&\geq& \langle\vct y,\vct\nu_\parallel\rangle - (3\epsilon+10\eta)\max(1,\|\vct\nu\|,\|\vct\nu\|^2,\|\vct\nu\|^3) +\langle\tilde{\vct z}',\tilde{\vct\nu}'\rangle - \frac{\lambda}{2}\|\tilde{\vct\nu}'\|_2^2\\
&\geq&  \langle\vct y,\vct\nu_\parallel\rangle - (3\epsilon+10\eta)\max(1,|\vct\nu\|^3) - \sqrt{\frac{1+\epsilon}{1-\epsilon}}\|\tilde{\vct z}\|_2\|\tilde{\vct\nu}'\|_2 - \frac{\lambda}{2}\cdot (1-\epsilon)\|\tilde{\vct\nu}\|^2\\
&\geq&  \langle\vct y,\vct\nu_\parallel\rangle - (3\epsilon+10\eta)\max(1,|\vct\nu\|^3) - 2\eta'\|\tilde{\vct\nu}'\| - \frac{\lambda}{2}\cdot (1-\epsilon)(1+\epsilon)\|\vct\nu_\parallel\|^2\\
&\geq& \langle\vct y,\vct\nu_\parallel\rangle - \frac{\lambda}{2}\|\vct\nu_\parallel\|^2 - (3\epsilon+10\eta)\max(1,\|\vct\nu\|^3) - 4\eta\|\tilde{\vct\nu}'\|\\
&\geq& f(\vct\nu_\parallel) -  (3\epsilon+13\eta)\max(1,\|\vct\nu\|^3)\\
&\geq& f(\vct\nu) - \frac{2\lambda\eta^2}{\rho_\ell} - (3\epsilon+13\eta)\max(1,\|\vct\nu\|^3).
\end{eqnarray*}
Similarly, one can show that
\begin{equation}
\tilde f(\vct\nu^*) \leq f(\vct\nu) + \frac{2\lambda\eta'^2}{\rho_\ell} + (3\epsilon+13\eta')\max(1,\|\vct\nu^*\|^3)
\leq f(\vct\nu) + \frac{3\lambda\eta^2}{\rho_\ell} + (3\epsilon+19\eta)\max(1,\|\vct\nu^*\|^3).
\end{equation}
Consequently, noting that $\tilde f(\tilde{\vct\nu}') \leq \tilde f(\vct\nu^*)$ one has
\begin{equation}
\big| \tilde f(\vct\nu^*) - \tilde f(\tilde{\vct\nu}')\big| \leq \frac{5\lambda\eta^2}{\rho_\ell} + (6\epsilon+32\eta)\max(1,\|\vct\nu\|^3,\|\vct\nu^*\|^3).
\end{equation}
Since both dual problems (before and after projection) are strongly convex with parameter $\lambda$, the following perturbation bound on $\|\vct\nu^*-\tilde{\vct\nu}'\|$ holds:
\begin{equation}
\|\vct\nu^* - \tilde{\vct\nu}'\| \leq \sqrt{\frac{2|\tilde f(\vct\nu^*) - \tilde f(\tilde{\vct\nu}')|}{\lambda}}
\leq \sqrt{\frac{10\eta^2}{\rho_\ell} + \frac{(12\epsilon+64\eta)\max(1,\|\vct\nu\|^3,\|\vct\nu^*\|^3)}{\lambda}}.
\end{equation}
Subsequently,
\begin{eqnarray*}
\|\tilde{\vct v}'-\vct v^*\|
&\leq& \frac{2\|\tilde{\vct\nu}' - \vct\nu^*\|}{\max(\|\tilde{\vct\nu}'\|, \|\vct\nu^*\|)}\\
&\leq& \frac{8\|\tilde{\vct\nu}' - \vct\nu^*\|}{\max(\|\vct\nu\|,\|\vct\nu^*\|)}\\
&\leq& 8\sqrt{\frac{10\eta^2}{\rho_\ell\max(\|\vct{\nu}\|^2,\|\vct\nu^*\|^2)} + \frac{12\epsilon+64\eta}{\lambda}\cdot \max\left(\|\vct\nu\|,\|\vct\nu^*\|,\frac{1}{\|\vct\nu\|^2},\frac{1}{\|\vct\nu^*\|^2}\right)}\\
&\leq& 8\sqrt{\frac{10\eta^2}{\rho_\ell(1-\lambda/2)^2} + \frac{12\epsilon+64\eta}{\lambda}\left(\frac{1}{(1-\lambda/2)^2} + \sqrt{\frac{2}{\lambda}}\right)}\\
&\leq& 16\sqrt{\frac{10\eta^2}{\rho_\ell} + \frac{12\epsilon+64\eta}{\lambda}}\\
&=& O\left(\sqrt{\frac{\eta^2}{\rho_\ell} + \frac{\epsilon+\eta}{\lambda}\sqrt{\frac{2}{\lambda}}}\right)\\
&=& O\left(\eta\rho_\ell^{-1/2} + (\epsilon+\eta)^{1/2}\lambda^{-3/4}\right).
\end{eqnarray*}
The last inequality is due to the fact that $\sqrt{a+b}\leq\sqrt{a}+\sqrt{b}$ for any $a,b\geq 0$.
Finally, the perturbation of the angle between $\vct v$ and $\vct y$ can be bounded by
\begin{multline*}
\big|\langle\vct v,\vct y\rangle - \langle\vct v^*, \tilde{\vct y}'\rangle\big| \leq \|\tilde{\vct v}'-\tilde{\vct v}^*\| + 2\epsilon
\leq O\left(\eta\rho_\ell^{-1/2} + (\epsilon+\eta)^{1/2}\lambda^{-3/4}\right) + 2\epsilon\\
= O\left(\eta\rho_\ell^{-1/2} + (\epsilon+\eta)^{1/2}\lambda^{-3/4}\right).
\end{multline*}

\end{proof}

With Lemma \ref{lem_subspace_incoherence_noisy} the following corollary on subspace incoherence perturbation immediately follows.
\begin{cor}[Perturbation of subspace incoherence under deterministic noise]
Assume the conditions as in Lemma \ref{lem_subspace_incoherence_noisy}.
Let $\mu_\ell$ and $\tilde\mu_\ell$ be the subspace incoherence before and after dimension reduction.
Then we have
\begin{equation}
\tilde\mu_\ell \leq \mu_\ell + O\left(\eta\rho_\ell^{-1/2} + (\epsilon+\eta)^{1/2}\lambda^{-3/4}\right),\quad\forall \ell\in [k].
\label{eq_perturbation_mu_noisy}
\end{equation}
\label{cor_perturbation_mu_noisy}
\end{cor}

We are now ready to prove Theorem \ref{thm_main_noisy} and \ref{thm_main_noisy_rand},
our main theorem stating deterministic success conditions for dimensionality-reduced Lasso SSC on data corrupted by adversarial or stochastic noise.

\begin{proof}[Proof of Theorem \ref{thm_main_noisy}]
Define $\tilde\Delta := \min_\ell(\tilde\rho_\ell-\tilde\mu_\ell)$ to be the maximum margin of error after dimensionality reduction.
%First we prove that with $\lambda=\rho/4 < 1/4$ and the upper bound in Eq. (\ref{eq_eps_noisy}) we have $\tilde\Delta \geq \Delta/2$,
We first prove that with $\epsilon$ upper bounded as in Eq. (\ref{eq_eps_noisy}), $\tilde\Delta$ satisfies $\tilde\Delta\geq\Delta/2$,
where $\Delta=\min_\ell(\rho_\ell-\mu_\ell)$ is the maximum margin of error before dimensionality reduction.
Essentially, this requires 
\begin{eqnarray*}
O\left(\eta\rho_\ell^{-1/2}\right) &<& \frac{\Delta}{4},\\
O\left((\epsilon+\eta)^{1/2}\lambda^{-3/4}\right) &<& \frac{\Delta}{4},
\end{eqnarray*}
which amounts to
\begin{eqnarray}
\eta &=& O\left(\Delta\cdot \lambda^{3/4}\right),\nonumber\\
\epsilon+\eta &=& O\left(\Delta^2\cdot\lambda^{3/2}\right). \label{eq_sum_bound}
\end{eqnarray}
Notice that the second bound implies the first one, up to numerical constants.

Next we verify that Eq. (\ref{eq_noisy_lambda_range}) is satisfied after dimensionality reduction.
By Eq. (\ref{eq_sum_bound}), we can safely assume that $\eta < \frac{\rho_\ell}{96}$ because $\Delta < \rho_\ell$.
Let $\tilde\eta$ denote the noise level after projection, that is, $\max_i\{\|\tilde{\vct z}_i\|\} \leq\tilde\eta$.
Because $\epsilon<1$, by Proposition \ref{prop_well_behave}
$\tilde\eta \leq 2\eta$ with high probability.
Consequently, $\eta < \frac{\rho}{96}$ implies ($\tilde\rho = \min_\ell{\tilde\rho_\ell}$ and $\tilde\mu=\max_\ell{\tilde\mu_\ell}$)
\begin{equation}
\tilde\rho - 2\tilde\eta - \tilde\eta^2 \geq \rho(1-\epsilon) - 6\eta  > \lambda.
\end{equation}
Hence the upper bound on $\lambda$ in Eq. (\ref{eq_noisy_lambda_range}) is satisfied.
For the lower bound, note that $\eta< 1/96 < 1$, $\tilde\rho_\ell<1$ and hence
\begin{equation}
\frac{\tilde\eta(1+\tilde\eta)(2+\tilde\rho_\ell)}{\tilde\rho_\ell-\tilde\mu_\ell-2\tilde\eta}
\leq \frac{6\tilde\eta}{\tilde\Delta} \leq \frac{12\eta}{\Delta/2} = \frac{24\eta}{\Delta} < \lambda.
\end{equation}
Therefore Eq. (\ref{eq_noisy_lambda_range}) is satisfied when $\lambda$ falls in the particular range, provided that $\epsilon$ and $\eta$
are bounded as in the statement of the theorem.
Finally, to satisfy $\epsilon+4\eta < \frac{1}{2\max(1,\|\vct\nu\|^2)}$ as in the conditions of Lemma \ref{lem_subspace_incoherence_noisy},
we need $\epsilon+4\eta < \lambda/4$ (because $\|\vct\nu\|^2\leq 2/\lambda$), which yields $\epsilon < \frac{\lambda}{4}-4\eta$.
This is clearly implied by the condition as in Eq. (\ref{eq_sum_bound}).
\end{proof}

\subsection{Proof of Theorem \ref{thm_main_noisy_rand}}

	We now proceed to prove Theorem \ref{thm_main_noisy_rand}, which shows that the dimensionality-reduced noisy sparse subspace clustering algorithm
	is capable of tolerating a significantly large amount of noise if the noise is stochastic (in particular, Gaussian) rather than adversarial.
	Similar to the case of Theorem \ref{thm_main_noisy}, the only argument that needs to be revised is the perturbation result for subspace incoherence.
	To exploit the stochasticity of the noise in the random noise model, we need to sharpen our bound for the adversarial noise in Lemma~\ref{lem_subspace_incoherence_noisy}. In particular, we can have $\epsilon$ to depend on $\xi\eta$ rather than $\eta$, where $\xi\asymp \sqrt{\frac{r\log N}{d}}$ goes to zero as the ambient dimension $d$ increases.
	In particular, we have the following lemma:
\begin{lem}[Perturbation of dual directions under stochastic noise]
Assume $0<\lambda<1$.
Suppose $\mat X=\mat Y+\mat Z$ where $\mat Y$ is the uncorrupted data matrix and $\mat Z\sim\nml(\vct 0,\frac{\sigma^2}{d}\mat I_{d\times N})$ is a random Gaussian noise matrix.
Fix a column $\vct x$ with dual direction $\vct\nu$ and $\vct v$ defined in Eq. (\ref{eq_lasso_dual}) and (\ref{eq_subspace_incoherence}).
Suppose $\widetilde{\mat Y} = \mat\Psi\mat Y$ is the projected noiseless data matrix and $\widetilde{\mat Y}'$ is the normalized version of $\widetilde{\mat Y}$.
Let $\widetilde{\mat X}' = \widetilde{\mat Y}'+\widetilde{\mat Z}'$ be the noisy observation after projection,
where $\widetilde{\mat Z}'$ is the normalized version of the projected noise matrix $\widetilde{\mat Z}=\mat\Psi\mat Z$.
Suppose $\|\mat\Psi\|_2 \leq \psi\sqrt{\frac{d}{p}}$ for some $\psi=O(1)$.
In addition, if $\mat\Psi$ satisfies Eq. (\ref{eq_well_behave},\ref{eq_well_behave_jl}) with parameter $\epsilon$
and $(\epsilon+\xi\eta) < 1/2\max(1,\|\vct\nu\|^2)$ then the following holds for all $\vct w\in\mat X\backslash\mat X^{(\ell)}$:
\begin{equation}
\big| \langle\vct v,\vct w\rangle - \langle\vct v^*,\tilde{\vct w}'\rangle\big| = O\left(\xi^{1/2}\sigma\rho_\ell^{-1/2}\log N + (\epsilon+\xi\sigma\log N)^{1/2}\lambda^{-3/4}\right),
\label{eq_perturbation_nu_stochastic}
\end{equation}
where $\xi=C\sqrt{\frac{r\log N}{d}}$ for some absolute constant $C$.
\label{lem_subspace_incoherence_stochastic}
\end{lem}

Before proving Lemma \ref{lem_subspace_incoherence_stochastic}, we first state a technical proposition that exploit the concentration properties
of Gaussian random vectors. 
These properties appear and are proved in \citep{noisy-ssc,robust-ssc}.
%\begin{prop}[Lemma 17, \cite{noisy-ssc}]
%Fix $\vct z\in\mathbb R^d$ and $\epsilon\in(0,1)$.
%Let $\vct a\in\mathbb R^d$ be a random vector sampled uniformly at random from the unit sphere in $\mathbb R^n$.
%Then we have
%\begin{equation}
%\Pr\left[\big|\vct a^\top\vct z\big| > \epsilon\right] \leq 2e^{-\frac{d\epsilon^2}{2}}.
%\label{eq_spherical_cap}
%\end{equation}
%\end{prop}

\begin{prop}[Lemma 18, \cite{noisy-ssc}]
Fix $\vct z\in\mathbb R^d$ and $\epsilon \in (0,1)$.
Suppose $\vct a\sim\nml(\vct 0, \frac{\sigma^2}{d}\mat I_{d\times d})$ is a $d$-dimensional Gaussian random vector.
Then we have
\begin{eqnarray}
\Pr\left[\|\vct a\|_2^2 > (1+\epsilon)\sigma^2\right] &\leq& e^{\frac{d}{2}(\log(1+\epsilon)-\epsilon)};\label{eq_gauss_norm}\\
\Pr\left[\big|\vct a^\top\vct z\big| > \epsilon\|\vct a\|\|\vct z\|\right] &\leq& 2e^{-\frac{n\epsilon^2}{2}}.\label{eq_gauss_inner_product}
\end{eqnarray}
\end{prop}

We are now ready to prove Lemma \ref{lem_subspace_incoherence_stochastic}.

\begin{proof}[Proof of Lemma \ref{lem_subspace_incoherence_stochastic}]

We first make some easy observations on the stochastic noise $\vct z$ and $\tilde{\vct z}=\mat\Psi\vct z$
before and after dimensionality reduction.
Since $\vct z\mapsto \mat\Psi\vct z$ is a linear mapping, we conclude that $\tilde{\vct z}$ is still a Gaussian random vector.
Because the theory of sparse subspace clustering is rotation invariant, we can properly rotate the space so that $\tilde{\vct z}$ has diagonal covariance matrix;
furthermore, the largest element-wise variance of $\tilde{\vct z}$ is upper bounded by $\psi^2\sigma^2/p$,
where $\psi$ is an upper bound on $\sqrt{\frac{p}{d}}\|\mat\Psi\|_2$ and is assumed to behave like a constant (i.e., $\psi=O(1)$).

Fix $\ell\in\{1,2,\cdots,k\}$ and a particular column $\vct x=\vct x_i\in\mathbb X\backslash\mathbb X^{(\ell)}$.
Suppose $\vct\nu$ is the optimal solution to the original dual problem in Eq. (\ref{eq_lasso_dual}).
Define $\vct\nu_{\parallel} = \mathcal P_{\mathcal U^{(\ell)}}\vct\nu$ and $\vct\nu_\perp = \mathcal P_{\mathcal U^{(\ell)^\perp}}\vct\nu$.
Let $f(\cdot)$ be the objective value of the dual problem under a specific solution.
%In addition, define $\xi = C\sqrt{\frac{r\log N}{d}}$ where $C$ is a numerical constant.
In addition, denote $\eta$ as the maximum magnitude of $\vct z_i$ (i.e., $\max_i\|\vct z_i\|_2\leq \eta$).
Then it is easy to observe that
\begin{eqnarray}
f(\vct\nu_\parallel)
&=& \langle\vct\nu_\parallel, \vct x\rangle - \frac{\lambda}{2}\|\vct\nu_\parallel\|_2^2\nonumber\\
&\geq& \langle\vct\nu_\parallel,\vct x\rangle - \frac{\lambda}{2}\|\vct\nu\|_2^2\nonumber\\
&=& \langle\vct\nu,\vct x\rangle - \langle\vct\nu_\perp,\vct x\rangle - \frac{\lambda}{2}\|\vct\nu\|_2^2\nonumber\\
 &=& f(\vct\nu) - \langle \vct x, \vct\nu_{\perp}\rangle \nonumber\\
 &\geq& f(\vct\nu) - \xi\eta\|\vct\nu_\perp\|_2,\nonumber
\end{eqnarray}
where in the last inequality we apply Eq. (\ref{eq_gauss_inner_product}) and the fact that $\vct z_i$ is independent from $\vct\nu_\perp$,
since $\vct z_i\in\mathcal X\backslash\mathcal X^{(\ell)}$ and $\vct\nu_\perp$ only depends on $\mathcal X^{(\ell)}$.
We then cite the following upper bound for $\|\vct\nu_\perp\|$, which appears as Eq. (5.16) in \citep{noisy-ssc}.
\begin{equation}
\|\vct\nu_\perp\|_2 \leq \lambda\eta\left(\frac{1}{r(\mathcal Q(\mat Y_{-i}^{(\ell)}))} + 1\right) \leq \frac{2\lambda\eta}{\rho_\ell}.
\label{upper_bound_nu2}
\end{equation}

Let $\tilde{\vct\nu} = \mat\Psi\vct\nu_\parallel$ and 
$$\tilde{\vct\nu}' = \frac{\sqrt{1-\epsilon}}{1+(\epsilon+(3+\psi^2)\xi\eta)\max(1,\|\vct\nu\|^2)}\cdot\tilde{\vct\nu}.$$
It is easy to verify that $\tilde{\vct\nu}'$ is a feasible solution to the projected dual problem.
To see this, note that
\begin{eqnarray*}
\langle\widetilde{\vct x}, \widetilde{\vct\nu}\rangle
&=& \langle\mat\Psi\vct x_\parallel, \mat\Psi\vct\nu_\parallel\rangle + \langle\mat\Psi\vct x_\perp,\mat\Psi\vct\nu_\parallel\rangle\\
&\leq& \langle\mat\Psi\vct x_\parallel, \mat\Psi\vct\nu_\parallel\rangle + (1+\epsilon)\xi\eta\|\vct\nu_\parallel\|\\
&=& \langle\mat\Psi\vct y,\mat\Psi\vct\nu_\parallel\rangle + \langle\mat\Psi\vct z_\parallel, \mat\Psi\vct\nu_\parallel\rangle + (1+\epsilon)\xi\eta\|\vct\nu\|\\
&\leq& \langle\mat\Psi\vct y,\mat\Psi\vct\nu_\parallel\rangle + \xi\eta\psi^2\|\vct\nu_\parallel\| + (1+\epsilon)\xi\eta\|\vct\nu\|\\
&\leq& \langle\vct y, \vct\nu_\parallel\rangle + \epsilon\cdot\frac{\|\vct y\|^2 + \|\vct\nu_\parallel\|^2}{2} + \xi\eta\psi^2\|\vct\nu_\parallel\| + (1+\epsilon)\xi\eta\|\vct\nu\|^2\\
&=& \langle\vct x,\vct\nu\rangle - \langle\vct z,\vct\nu\rangle + \epsilon\max(1,\|\vct\nu\|^2) + \xi\eta\psi^2\|\vct\nu_\parallel\| + (1+\epsilon)\xi\eta\|\vct\nu\|^2\\
&\leq& 1 + \xi\eta\|\vct\nu\| + \epsilon\max(1,\|\vct\nu\|^2) + \xi\eta\psi^2\|\vct\nu_\parallel\| + (1+\epsilon)\xi\eta\|\vct\nu\|^2\\
&\leq& 1 + (\epsilon+(3+\psi^2)\xi\eta)\max(1,\|\vct\nu\|^2).
%&\leq& 1 + \eta\|\vct\nu_\perp\| + \epsilon\max(1,\|\vct\nu\|^2) + \epsilon\eta + (1+\epsilon)\xi\eta\|\vct\nu\|\\
%&\leq& 1 + \epsilon\eta + (2+\epsilon)\eta\|\vct\nu\| + \epsilon\max(1,\|\vct\nu\|^2)\\
%&\leq& 1 + (\epsilon+4\eta)\max(1,\|\vct\nu\|,\|\vct\nu\|^2)\\
%&\leq& 1 + (\epsilon+4\eta)\max(1,\|\vct\nu\|^2).
\end{eqnarray*}
%In the above derivation, for the third and fourth lines we use the fact that $\mat\Psi$ is well behaved and $\vct x_\perp=\vct z$, $\vct\nu_\parallel\in\mathcal U^{(\ell)}$.
In the above derivation, for the fourth line we use the fact that $\mat\Psi\vct z$ is a random Gaussian vector independent of $\mat\Psi\vct\nu$ (conditioned on $\mat\Psi$) as discussed above
and the assumption that $\|\mat\Psi\|_2\leq\psi$.
For the fifth line we use the condition that $\mat\Psi$ is well-behaved and apply Proposition \ref{prop_well_behave}.
To see why the sixth line holds, notice that
$$
\langle\vct x,\vct\nu\rangle = \langle\vct y+\vct z,\vct\nu_\parallel+\vct\nu_\perp\rangle = \langle\vct y,\vct\nu_\parallel\rangle + \langle\vct z,\vct\nu\rangle
$$
because $\langle\vct y,\vct\nu_\perp\rangle = 0$.

With the upper bound on $\langle\widetilde{\vct x},\widetilde{\vct\nu}\rangle$, we can subsequently upper bound $\|\widetilde{\mat X}'^\top\widetilde{\vct\nu}'\|_{\infty}$ as
$$
\|\widetilde{\mat X}'^\top\widetilde{\vct\nu}'\|_{\infty}
\leq \frac{1}{\sqrt{1-\epsilon}}\cdot \frac{\sqrt{1-\epsilon}}{1+(\epsilon+(3+\psi^2)\xi\eta)\max(1,\|\vct\nu\|^2)}\cdot \|\widetilde{\mat X}^\top\widetilde{\vct\nu}\|_{\infty} \leq 1.
$$

%{\color{red}
%Next,
%define $\eta' := \max_{i=1,\cdots,n}\|\tilde{\vct z}_i\|_2$.
%Since $\mat\Psi$ is well behaved, $\eta' \leq \sqrt{1+\epsilon}\eta$ with high probability.}
We now proceed to lower bound $\tilde f(\tilde{\vct\nu}')$.
First note that $\langle\tilde{\vct y},\tilde{\vct \nu}\rangle\geq 0$ with high probability because
\begin{eqnarray*}
\langle\tilde{\vct y},\tilde{\vct\nu}\rangle 
&\geq& \langle\vct y,\vct\nu\rangle - \epsilon\cdot \frac{\|\vct y\|^2 + \|\vct\nu\|^2}{2}\\
&=& \langle\vct x,\vct\nu\rangle - \epsilon\cdot\frac{\|\vct y\|^2 + \|\vct\nu\|^2}{2} - \langle\vct z,\vct\nu\rangle\\
&\geq& \langle\vct x,\vct\nu\rangle - \epsilon\max(1,\|\vct\nu\|^2) - \xi\eta\|\vct\nu\|\\
&\geq& 0,
\end{eqnarray*}
where in the last inequality we use the fact that $\langle\vct x,\vct\nu\rangle\geq 1-\lambda/2$ and the assumption that $\lambda\in(0,1)$ and $(\epsilon+\xi\eta)\max(1,\|\vct\nu\|^2) < 1/2$.
Applying essentially the same chain of argument as in the proof of Lemma \ref{lem_subspace_incoherence_noiseless}
we obtain
\begin{eqnarray*}
\tilde f(\tilde{\vct\nu}')
&=& \langle\tilde{\vct x}', \tilde{\vct\nu}'\rangle - \frac{\lambda}{2}\|\tilde{\vct\nu}'\|_2^2\\
&=& \langle\tilde{\vct y}', \tilde{\vct\nu}'\rangle + \langle\tilde{\vct z}', \tilde{\vct\nu}'\rangle - \frac{\lambda}{2}\|\tilde{\vct\nu}'\|_2^2\\
&\geq& \sqrt{\frac{1-\epsilon}{1+\epsilon}}\frac{\langle\tilde{\vct y},\tilde{\vct\nu}\rangle}{1+(\epsilon+(3+\psi^2)\xi\eta)\max(1,\|\vct\nu\|^2)} + \langle\tilde{\vct z}',\tilde{\vct\nu}'\rangle - \frac{\lambda}{2}\|\tilde{\vct\nu}'\|_2^2\\
&\geq& (1-\epsilon)(1-(\epsilon+(3+\psi^2)\xi\eta)\max(1,\|\vct\nu\|^2))(\langle\vct y,\vct\nu_\parallel\rangle - (\epsilon+(3+\psi^2)\xi\eta)\max(1,\|\vct\nu\|^2))\\
&& +\langle\tilde{\vct z}',\tilde{\vct\nu}'\rangle - \frac{\lambda}{2}\|\tilde{\vct\nu}'\|_2^2\\
&\geq& (1-\epsilon)(\langle\vct y,\vct\nu_\parallel\rangle - (\epsilon+(3+\psi^2)\xi\eta)\max(1,\|\vct\nu\|^2)) - (\epsilon+(3+\psi^2)\xi\eta)\max(1,\|\vct\nu\|^2)\cdot\langle\vct y,\vct\nu_\parallel\rangle\\
&& +\langle\tilde{\vct z}',\tilde{\vct\nu}'\rangle - \frac{\lambda}{2}\|\tilde{\vct\nu}'\|_2^2\\
&\geq& \langle\vct y,\vct\nu_\parallel\rangle - \epsilon\|\vct\nu_\parallel\| - (\epsilon+(3+\psi^2)\xi\eta)\max(1,\|\vct\nu\|^2) - (\epsilon+(3+\psi^2)\xi\eta)\max(1,\|\vct\nu\|^2)\cdot\langle\vct y,\vct\nu_\parallel\rangle\\
&& +\langle\tilde{\vct z}',\tilde{\vct\nu}'\rangle - \frac{\lambda}{2}\|\tilde{\vct\nu}'\|_2^2\\
&\geq& \langle\vct y,\vct\nu_\parallel\rangle - (3\epsilon+(6+2\psi^2)\xi\eta)\max(1,\|\vct\nu\|,\|\vct\nu\|^2,\|\vct\nu\|^3) +\langle\tilde{\vct z}',\tilde{\vct\nu}'\rangle - \frac{\lambda}{2}\|\tilde{\vct\nu}'\|_2^2\\
&\geq&  \langle\vct y,\vct\nu_\parallel\rangle - (3\epsilon+(6+2\psi^2)\xi\eta)\max(1,|\vct\nu\|^3) - \frac{\psi}{\sqrt{1-\epsilon}}\xi\eta\|\tilde{\vct\nu}'\|_2 - \frac{\lambda}{2}\cdot (1-\epsilon)\|\tilde{\vct\nu}\|^2\\
&\geq&  \langle\vct y,\vct\nu_\parallel\rangle - (3\epsilon+(6+2\psi^2)\xi\eta)\max(1,|\vct\nu\|^3) - \psi\xi\eta\|\tilde{\vct\nu}\|_2 - \frac{\lambda}{2}\cdot (1-\epsilon)(1+\epsilon)\|\vct\nu_\parallel\|^2\\
&\geq& \langle\vct y,\vct\nu_\parallel\rangle - \frac{\lambda}{2}\|\vct\nu_\parallel\|^2 - (3\epsilon+(6+2\psi^2)\xi\eta)\max(1,\|\vct\nu\|^3) - \psi^2\xi\eta\|\vct\nu\|\\
&\geq& f(\vct\nu_\parallel) -  (3\epsilon+(6+3\psi^2)\xi\eta)\max(1,\|\vct\nu\|^3)\\
&\geq& f(\vct\nu) - \frac{2\lambda\xi\eta^2}{\rho_\ell} - (3\epsilon + (6+3\psi^2)\xi\eta)\max(1,\|\vct\nu\|^3).
\end{eqnarray*}
Similarly, denoting $\eta'$ as $\eta'=\max_i{\|\tilde{\vct z}_i'\|}$ one can show that
\begin{multline}
\tilde f(\vct\nu^*) \leq f(\vct\nu) + \frac{2\lambda\xi\eta'^2}{\rho_\ell} + (3\epsilon+(6+3\psi^2)\xi\eta')\max(1,\|\vct\nu^*\|^3)\\
\leq f(\vct\nu) + \frac{2\lambda\psi^2\xi\eta^2}{\rho_\ell} + (3\epsilon+(6+3\psi^2)\psi^2\xi\eta)\max(1,\|\vct\nu^*\|^3).
\end{multline}
Consequently, noting that $\tilde f(\tilde{\vct\nu}') \leq \tilde f(\vct\nu^*)$ one has
\begin{equation}
\big| \tilde f(\vct\nu^*) - \tilde f(\tilde{\vct\nu}')\big| \leq \frac{4\lambda\psi^2\xi\eta^2}{\rho_\ell} + (6\epsilon+18\psi^4\xi\eta)\max(1,\|\vct\nu\|^3,\|\vct\nu^*\|^3).
\end{equation}
Since both dual problems (before and after projection) are strongly convex with parameter $\lambda$, the following perturbation bound on $\|\vct\nu^*-\tilde{\vct\nu}'\|$ holds:
\begin{equation}
\|\vct\nu^* - \tilde{\vct\nu}'\| \leq \sqrt{\frac{2|\tilde f(\vct\nu^*) - \tilde f(\tilde{\vct\nu}')|}{\lambda}}
\leq \sqrt{\frac{8\psi^2\xi\eta^2}{\rho_\ell} + \frac{(12\epsilon+36\psi^4\xi\eta)\max(1,\|\vct\nu\|^3,\|\vct\nu^*\|^3)}{\lambda}}.
\end{equation}
Subsequently,
\begin{eqnarray*}
\|\tilde{\vct v}'-\vct v^*\|
&\leq& \frac{2\|\tilde{\vct\nu}' - \vct\nu^*\|}{\max(\|\tilde{\vct\nu}'\|, \|\vct\nu^*\|)}\\
&\leq& \frac{8\|\tilde{\vct\nu}' - \vct\nu^*\|}{\max(\|\vct\nu\|,\|\vct\nu^*\|)}\\
&\leq& 8\sqrt{\frac{8\psi^2\xi\eta^2}{\rho_\ell\max(\|\vct{\nu}\|^2,\|\vct\nu^*\|^2)} + \frac{12\epsilon+36\psi^4\xi\eta}{\lambda}\cdot \max\left(\|\vct\nu\|,\|\vct\nu^*\|,\frac{1}{\|\vct\nu\|^2},\frac{1}{\|\vct\nu^*\|^2}\right)}\\
&\leq& 8\sqrt{\frac{8\psi^2\xi\eta^2}{\rho_\ell(1-\lambda/2)^2} + \frac{12\epsilon+36\psi^4\xi\eta}{\lambda}\left(\frac{1}{(1-\lambda/2)^2} + \sqrt{\frac{2}{\lambda}}\right)}\\
&\leq& 16\sqrt{\frac{8\psi^2\xi\eta^2}{\rho_\ell} + \frac{12\epsilon+36\psi^4\xi\eta}{\lambda}}\\
&=& O\left(\sqrt{\frac{\xi\eta^2}{\rho_\ell} + \frac{\epsilon+\xi\eta}{\lambda}\sqrt{\frac{2}{\lambda}}}\right)\\
&=& O\left(\xi^{1/2}\eta\rho_\ell^{-1/2} + (\epsilon+\xi\eta)^{1/2}\lambda^{-3/4}\right).
\end{eqnarray*}
The last inequality is due to the fact that $\sqrt{a+b}\leq\sqrt{a}+\sqrt{b}$ for any $a,b\geq 0$.
Finally, the perturbation of the angle between $\vct v$ and $\vct y$ can be bounded by
\begin{multline*}
\big|\langle\vct v,\vct y\rangle - \langle\vct v^*, \tilde{\vct y}'\rangle\big| \leq \|\tilde{\vct v}'-\tilde{\vct v}^*\| + 2\epsilon
\leq O\left(\xi^{1/2}\eta\rho_\ell^{-1/2} + (\epsilon+\xi\eta)^{1/2}\lambda^{-3/4}\right) + 2\epsilon\\
= O\left(\xi^{1/2}\eta\rho_\ell^{-1/2} + (\epsilon+\xi\eta)^{1/2}\lambda^{-3/4}\right)
= O\left(\xi^{1/2}\sigma\rho_\ell^{-1/2}\log N + (\epsilon + \xi\sigma\log N)^{1/2}\lambda^{-3/4}\right).
\end{multline*}

\end{proof}		

The following corollary immediately follows Lemma \ref{lem_subspace_incoherence_stochastic}.

\begin{cor}[Perturbation of subspace incoherence under stochastic noise]
Assume the conditions as in Lemma \ref{lem_subspace_incoherence_stochastic}.
Let $\mu_\ell$ and $\tilde\mu_\ell$ be the subspace incoherence before and after dimension reduction.
Then we have
\begin{equation}
\tilde\mu_\ell \leq \mu_\ell + O\left(\xi^{1/2}\sigma\rho_\ell^{-1/2}\log N + (\epsilon+\xi\sigma\log N)^{1/2}\lambda^{-3/4}\right),\quad\forall \ell\in [k].
\label{eq_perturbation_mu_stochastic}
\end{equation}
\label{cor_perturbation_mu_stochastic}
\end{cor}

We are now ready to prove Theorem \ref{thm_main_noisy_rand}.

\begin{proof}[Proof of Theorem \ref{thm_main_noisy_rand}]
The theorem would hold if the following inequalities are satisfied:
\begin{eqnarray}
\tilde\rho_\ell - \tilde\mu_\ell &>& \frac{\rho_\ell-\mu_\ell}{2};\label{eq_delta_perm}\\
\frac{\sigma(1+\sigma)}{\Delta}\sqrt{\frac{\log N}{d}} &<& \frac{\lambda}{100C_1C_2}.\label{eq_sigma_perm}
\end{eqnarray}
Here Eq. (\ref{eq_sigma_perm}) is due to Eq. (\ref{eq_sigma_lambda_bound}) and Eq. (\ref{eq_delta_perm}) ensures that we can safely
substitute $(\tilde\rho_\ell-\tilde\mu_\ell)$ with $(\rho_\ell-\mu_\ell)$ by incurring only a constant multiplicative term.
To satisfy Eq. (\ref{eq_delta_perm}), we may take $\xi^{1/2}\sigma\rho_\ell^{-1/2}\log N=O(\Delta)$ and
$\max(\epsilon,\xi\sigma\log N)=O(\Delta^2\lambda^{3/2})$,
which subsequently yields
\begin{equation}
\epsilon = O\left(\Delta^2\lambda^{3/2}\right);\quad
\sigma = O\left(\min\left\{\frac{\Delta\rho^{1/2}d^{1/4}}{r^{1/4}\log^{5/4}N}, \frac{\Delta^2\lambda^{3/2}d^{1/2}}{r^{1/2}\log^{3/2} N}\right\}\right).
\label{eq_condition_1_rand}
\end{equation}
On the other hand, Eq. (\ref{eq_sigma_perm}) implies
$$
\sigma = O\left(\max\left\{\frac{\Delta^{1/2}\lambda^{1/2}d^{1/4}}{\log^{1/4}N}, \min\left(1,\frac{\lambda\Delta d^{1/2}}{\log^{1/2}N}\right)\right\}\right),
$$
%which is superseded by the right-hand terms in Eq. (\ref{eq_condition_1_rand}) because $\Delta$ is within the range of $(0,1)$ and $\lambda\asymp \rho$.
%As a result, conditions in Eq. (\ref{eq_condition_1_rand}) is sufficient to guarantee subspace detection property of Lasso SSC under dimensionality reduction
%with stochastic noise.
\end{proof}

\subsection{Proof of Theorem \ref{thm_main_semirandom}}

In this section we prove Theorem \ref{thm_main_semirandom} which characterizes success conditions for dimensionality-reduced Lasso SSC under a semi-random data model.
The main tool for our proof is the following two theorems extracted from \cite{noisy-ssc}, 
which bound the subspace incoherence and inradius using affinity between subspaces.

\begin{lem}[Subspace incoherence bound under semi-random model; Lemma 22, \citep{noisy-ssc}]
Fix $t>0$.
In the semi-random model we have
\begin{equation}
\Pr\left[\forall \ell\neq\ell', \mu_\ell < t(\log((N_\ell+1)N_{\ell'})+\log k)\cdot \affn(\mathcal U^{(\ell)}, \mathcal U^{(\ell')})\right]
\geq 1 - \frac{1}{k^2}\sum_{\ell\neq \ell'}{\frac{1}{(N_\ell+1)N_{\ell'}}e^{-\frac{t}{4}}}.
\label{eq_semirandom_incoherence}
\end{equation}
\label{lem_semirandom_incoherence}
\end{lem}
	
\begin{lem}[Inradius bound under semi-random model; Lemma 21, \citep{noisy-ssc}]
Fix $\beta\in(0,1)$. Suppose for each subspace $\mathcal U^{(\ell)}$ we observe $N_\ell$ data points drawn uniformly at random from the unit sphere in $\mathcal U^{(\ell)}$
and $N_\ell\geq \kappa r_\ell$ for some $\kappa_\ell > 0$.
We then have
\begin{equation}
\Pr\left[\forall \ell, \rho_\ell \geq C\sqrt{\frac{\beta\log\kappa}{r_\ell}}\right] \geq 1-\sum_{\ell=1}^k{N_\ell e^{-r_\ell^\beta N_\ell^{1-\beta}}},
\label{eq_semirandom_inradius}
\end{equation}
where $C>0$ is an absolute constant.
\label{lem_semirandom_inradius}
\end{lem}

We are now ready to prove Theorem \ref{thm_main_semirandom} by plugging bounds in Lemma \ref{lem_semirandom_incoherence} and \ref{lem_semirandom_inradius}
into the perturbation analysis we derived in previous sections.
\begin{proof}[Proof of Theorem \ref{thm_main_semirandom}]
Setting $t\asymp\log N$ in Eq. (\ref{eq_semirandom_incoherence}) and $\beta=0.5$ in Eq. (\ref{eq_semirandom_inradius}), 
the following holds for all $\ell\neq \ell'$ with probability at least $1-O(1/N)$:
\begin{eqnarray}
\mu_\ell &\leq& O\left(\log^2(kN)\cdot \affn(\mathcal U^{(\ell)}, \mathcal U^{(\ell')})\right);\label{eq_semirandom_proof_eq1}\\
\rho_\ell &\geq& \Omega\left(\sqrt{\frac{\log\kappa}{r}}\right).\label{eq_semirandom_proof_eq2}
\end{eqnarray}
For Lasso SSC to hold on noisy data, we shall require $\mu_\ell < \rho_\ell / 10$.
\footnote{The constant 10 is unimportant and can be replaced with any other constant that is strictly larger than 1.}
By Eq. (\ref{eq_semirandom_proof_eq1}) and (\ref{eq_semirandom_proof_eq2}), a sufficient condition of $\mu_\ell < \rho_\ell / 10$ would be
$$
\min_{\ell\neq\ell'}\affn(\mathcal U^{(\ell)}, \mathcal U^{(\ell')}) = O\left(\frac{1}{\log^2(kn)}\sqrt{\frac{\log\kappa}{r}}\right),
$$
which is precisely the condition in Eq. (\ref{eq_semirandom_affinity}), as stated in Theorem \ref{thm_main_semirandom}.
Finally, by Theorem \ref{thm_main_noisy_rand} and note that
$$
\lambda\asymp\rho\asymp \sqrt{\frac{\log\kappa}{r}}
$$
we obtain the conditions on $\sigma$ and $\epsilon$ in Theorem \ref{thm_main_semirandom}.
That last equation, Eq. (\ref{eq_main_noisy_semirandom_p}), is a direct application of the bounds on $\epsilon$ and subspace embedding characterization of Gaussian random projections
in Proposition \ref{prop_wellbehave_jl}, which roughly states that
$$
p = \Omega\left(\frac{r\log(kN)}{\epsilon^2}\right)
$$
with probability $\geq 1-O(1/N)$.
\end{proof}

\subsection{Proof of Theorem \ref{thm_main_fullyrandom}}

Theorem \ref{thm_main_fullyrandom} follows immediate from the subsequent lemma that bounds subspace incoherence under the fully random model.
\begin{lem}[Subspace incoherence bound under fully random model;{\citealp[Lemma 23]{noisy-ssc}}]
Suppose $\mathcal U^{(1)},\cdots,\mathcal U^{(k)}$ are i.i.d.~drawn uniformly at random from all $r$-dimensional linear subspaces in a $d$-dimensional ambient space.
We then have
\begin{equation}
\Pr\left[\forall \ell, \mu_\ell \leq \sqrt{\frac{6\log N}{d}}\right] \geq 1 - \frac{2}{N}.
\end{equation}
\end{lem}

\subsection{Proof of Theorem \ref{thm:privacy}}\label{subsec:proof-privacy}

\begin{proof}
 Let $\mat X$ and $\mat X'$ differs by only one entry, w.l.o.g, assume it is the $i$th column and $j$th row,
$$\|\mat\Psi(\mat X-\mat X')\|_{F} =  \|\mat\Psi(\mat X_i-\mat X_i')\|_2 \leq  \|\mat\Psi \vct{e}_j\||\mat X_{ji}-\mat X_{ji}'| \leq 2\sqrt{\frac{\mu}{d}}\|\mat\Psi \vct{e}_j\|.$$
Now we derive the $\ell_2$-sensitivity of $\text{Normalize}(\mat\Psi(\cdot))$,
which is the $\ell_2$ distance between the vectorized output on any two neighboring input databases $\mat X$ and $\mat X'$ that
differ by only one entry:
\begin{eqnarray*}
&&\|\text{Normalize}(\mat\Psi\mat X) - \text{Normalize}(\mat\Psi\mat X')\|_{F} \\
&=& \left\| \frac{\mat\Psi\mat X_i}{\|\mat\Psi\mat X_i\|} - \frac{\mat\Psi\mat X_i'}{\|\mat\Psi\mat X_i'\|}\right\|_2 = \left\| \frac{\mat\Psi\mat X_i}{\|\mat\Psi\mat X_i\|} - \frac{\mat\Psi\mat X_i'}{\|\mat\Psi\mat X_i\|} + \frac{\mat\Psi\mat X_i'}{\|\mat\Psi\mat X_i\|}- \frac{\mat\Psi\mat X_i'}{\|\mat\Psi\mat X_i'\|}\right\|_2\\
&=&\left\| \frac{\mat\Psi (\mat X_i-\mat X_i')}{\|\mat\Psi\mat X_i\|} +\mat\Psi\mat X_i'\left(\frac{1}{\|\mat\Psi\mat X_i\|}-\frac{1}{\|\mat\Psi\mat X_i'\|}\right) \right\|\\
&\leq&  \frac{\|\mat\Psi(\mat X_i-\mat X_i')\|_2}{\|\mat\Psi\mat X_i\|}  + \|\mat\Psi\mat X_i'\| \frac{\left|\|\mat\Psi\mat X_i'\|-\|\mat\Psi\mat X_i\|\right|}{\|\mat\Psi\mat X_i\|\|\mat\Psi\mat X_i'\|}\\
&\leq& \frac{2\|\mat\Psi(\mat X_i-\mat X_i')\|_2}{\|\mat\Psi\mat X_i\|}\leq 4\sqrt{\frac{\mu_0}{d}}\frac{\|\mat\Psi \vct{e}_j\|}{\|\mat\Psi\mat X_i\|} \leq 4\sqrt{\frac{\mu_0}{d}}\frac{1+\epsilon}{1-\epsilon}.
\end{eqnarray*}
The last step uses the fact that $\mat\Psi$ is JL with parameter $\epsilon$.
\begin{lem}[Gaussian Mechanism, \citep{data-privacy-jl}]\label{lemma:gauss_mech}
Let $\Delta_2 f$ be the $\ell_2$ sensitivity of $f$,
Let $\varepsilon\in(0,1)$ be arbitrary. The procedure that output $f(\mat X) + \cN(0, \sigma^2 I)$ with $\sigma \geq \Delta_2f\sqrt{2\log(1.25/\delta)}/\varepsilon$ is $(\varepsilon,\delta)$-differentially private.
\end{lem}
Finally, the normalization step does not change privacy claim because differential privacy is close to post-processing.
\end{proof}

\subsection{Proof of Corollary \ref{cor:utility}}\label{subsec:proof-utility}

\begin{proof}
The proof involves applying Theorem~\ref{thm_main_noisy_rand} with $\xi=1$ and
$$\sigma=\frac{1+\epsilon}{1-\epsilon}\sqrt{\frac{32p\mu_0\log(1.25/\delta)}{d\varepsilon^2}} \leq \sqrt{\frac{128p\mu_0\log(1.25/\delta)}{d\varepsilon^2}}$$
according to Theorem~\ref{thm:privacy} and rearranging the expressions in terms of the limit for privacy requirement $\varepsilon$.

Note that the noise here is added after the compression and normalization, but the effect is the same as adding Gaussian noise in the original dimension and scaled orthogonal random projection on a noise. %In fact, we can replace $C/4$ with $C$ because there is no renormalization here.

Denote $B:=\min_{\ell=1,...,k}\{\rho, r^{-1/2}, \rho_\ell-\mu_\ell\}$.
 Theorem~\ref{thm_main_noisy_rand} shows that Lasso SSC succeeds if $\sigma$ satisfies
\begin{equation}\label{eq:utilityproof_eq}
\sigma = O\left(\min\left\{\frac{\log^{5/4}N}{B^2p^{1/4}}, \frac{\log^{3/2}N}{B^{11/2}p^{1/2}}\right\}\right).
\end{equation}
Substitute the expression of $\sigma$ into \eqref{eq:utilityproof_eq} and rewrite it in terms of $\varepsilon$, we get the lower bound of $\varepsilon$ as claimed in the statement of the corollary.
\end{proof}

\subsection{Proof of Proposition \ref{prop:impossible}}\label{subsec:proof-impossible}

\begin{proof}
First of all, if a data point can be arbitrarily replaced with another vector, then we can change it such that it goes into a different subspace.
Let's first ignore the gap between subspace detection property and perfect clustering. Assume that the output is the clustering result and it is always correct, then if we arbitrarily change the $k$th data point from one Subspace A to Subspace B, the result must reflect the change and cluster this data point correctly to its new subspace and the probability of observing an output that has $k$th data point clustered into Subspace A will change from $1$ to $0$, which blatantly violates the definition of differential privacy.

The same line of arguments holds if we treat the output as the graph embedding. Note that having subspace detection property (SDP) for data point $k$ in Subspace A (connected only to a set of points) and having subspace detection for data point $k$ in Subspace B (connected only to another set of points) are two disjoint measurable events. With a perturbation that changes a data point from one subspace to another will blow the likelihood ratio of observing one of these two event to infinity.

The high probability statement holds because 
$$\frac{\Pr( \text{SDP according to $\mat X$}|\mat X)}{\Pr(\text{SDP according to $\mat X$} |\mat X')} \geq  \frac{1-\delta}{\delta}\geq e^{\log(\frac{1-\delta}{\delta})}.$$
\end{proof}

\section{Conclusion}

We present theoretical analysis of Lasso SSC, one of the most popular algorithms for subspace clustering,
when data dimension is compressed due to resource constraints.
Our analysis applies to both deterministic and stochastic subspace data models and is capable of tolerating stochastic or even adversarial noise.
One interesting future direction is to further sharpen the dependence over $r$ (intrinsic dimension) in the lower bound of $p$ (dimension after projection) for Lasso SSC:
in our analysis (Theorems \ref{thm_main_semirandom}, \ref{thm_main_fullyrandom}) $p$ needs to be at least a low-degree polynomial of $r$, while an obvious lower bound for $p$ is $\Omega(r)$.
We conjecture that $p=\Omega(r\log N)$ is sufficient to guarantee SEP for Lasso SSC under random Gaussian projections (or any other projection that satisfies similar subspace embedding properties);
however, a rigorous proof might require substantially different techniques with regularization-independent perturbation analysis.

\section*{Acknowledgment}
This research is supported in part by grants NSF CAREER IIS-1252412 and AFOSR YIP FA9550-14-1-0285.
Yu-Xiang Wang was supported by NSF Award BCS-0941518 to CMU Statistics and Singapore National Research Foundation under its International Research Centre @ Singapore Funding Initiative and administered by the IDM Programme Office.

\begin{appendices}

%\section{Other proofs}\label{appsec:proof_ose}

\section{Technical proofs on subspace embedding properties}\label{appsec:proof_ose}

\begin{proof}[Proof of Proposition \ref{prop_well_behave}]
Fix $\ell,\ell'\in\{1,\cdots,k\}$ and let $\mathcal U = \text{span}(\mathcal U^{(\ell)}\cup \mathcal U^{(\ell')})$ denote the subspace
spanned by the union of the two subspaces $\mathcal U^{(\ell)}$ and $\mathcal U^{(\ell')}$.
By assumption, the rank of $\mathcal U^{(\ell)}\cup\mathcal U^{(\ell')}$, $r'$,  satisfies
$r'\leq r_{\ell}+r_{\ell'}\leq 2r$.
For any $\vct x\in\mathcal U^{(\ell)}$ and $\vct y\in\mathcal U^{(\ell')}$ we have
\begin{equation}
\langle\vct x,\vct y\rangle = \frac{1}{4}\left(\|\vct x+\vct y\|_2^2 - \|\vct x-\vct y\|_2^2\right);
\end{equation}
subsequently,
\begin{equation}
\big|\langle\vct x,\vct y\rangle - \langle\mat\Psi\vct x,\mat\Psi\vct y\rangle\big| \leq
\frac{1}{4}\left(\big|\|\vct x+\vct y\|^2 - \|\mat\Psi(\vct x+\vct y)\|^2\big| + \big|\|\vct x-\vct y\|^2 - \|\mat\Psi(\vct x-\vct y)\|^2\big|\right).
\label{eq_angle_norm}
\end{equation}
Since $\mat\Psi$ is a subspace embedding, the following holds for all $\vct x+\vct y,\vct x+\vct y\in\mathrm{span}(\mathcal U^{(\ell)}\cup\mathcal U^{(\ell')})$:
\begin{eqnarray*}
(1-\epsilon)^2\|\vct x+\vct y\|^2 \leq \|\mat\Psi(\vct x+\vct y)\|^2 \leq (1+\epsilon)^2\|\vct x+\vct y\|^2,\\
(1-\epsilon)^2\|\vct x-\vct y\|^2 \leq \|\mat\Psi(\vct x-\vct y)\|^2 \leq (1+\epsilon)^2\|\vct x-\vct y\|^2.
\end{eqnarray*}
The bound for $\big|\langle\vct x,\vct y\rangle - \langle\mat\Psi\vct x,\mat\Psi\vct y\rangle\big| $ then follows by noting that $(1-\epsilon)^2\geq 1-3\epsilon$, $(1+\epsilon)^2\leq 1+3\epsilon$
and $\|\vct x+\vct y\|^2 + \|\vct x-\vct y\|^2 = 2(\|\vct x\|^2+\|\vct y\|^2)$.
Finally, a union bound over all $k^2$ subspaces and $2N$ data points yields the proposition.
\end{proof}

\begin{proof}[Proof of Proposition \ref{prop_wellbehave_jl}]
Fix $\mathcal U\subseteq\mathbb R^d$ to be any subspace of dimension at most $r'$
and let $\mat U\in\mathbb R^{d\times r}$ be an orthonormal basis of $\mathcal U$.
Let $\widetilde{\mat\Psi} = \sqrt{p}\mat\Psi$ denote the unnormalized version of $\mat\Psi$.
Since each entry in $\widetilde{\mat\Psi}$ follows i.i.d. standard Gaussian distribution and $\mat U$ is orthogonal,
the projected matrix $\widetilde{\mat\Psi}\mat U\in\mathbb R^{p\times r'}$ follows an entrywise standard Gaussian distribution, too.
By Lemma~\ref{lem_gaussian_spectrum} (taking $t=\sqrt{2\delta}$ and scale the matrix by $1/\sqrt{p}$), the singular values of the Gaussian random matrix $\mat\Psi$ obey
\begin{equation}\label{eq:psi_specturm}
1-\sqrt{\frac{r'}{p}} - \sqrt{\frac{2\log(1/\delta)}{p}}\leq \sigma_{r'}(\mat\Psi)\leq \sigma_1(\mat\Psi)\leq 1+\sqrt{\frac{r'}{p}}+\sqrt{\frac{2\log(1/\delta)}{p}}
\end{equation}
with probability at least $1-\delta$. Let $\epsilon:=\sqrt{\frac{r'}{p}}+\sqrt{\frac{2\log(1/\delta)}{p}}$, then with the same probability,
(supposing $\vct x = \mat U\vct\alpha\in\mathcal U$)
\begin{eqnarray}
\big|\|\vct x\|_2^2 - \|\vct\Psi\vct x\|_2^2\big|
&=& \big| \vct\alpha^\top\mat U^\top\mat U\vct\alpha - \vct\alpha^\top\mat U^\top\mat\Psi^\top\mat\Psi\mat U\vct\alpha\big|  \nonumber\\
&\leq& \|\vct\alpha\|_2^2\|\mat U^\top\mat U-\mat U^\top\mat\Psi^\top\mat\Psi\mat U\|_2\nonumber\\
&=& \|\vct x\|_2^2\|\mat I_{r'\times r'} - \mat U^\top\mat\Psi^\top\mat\Psi\mat U\|_2\nonumber\\
&\leq& \epsilon\|\vct x\|_2^2. \label{eq_gaussian_norm_preservation}
\end{eqnarray}
Subsequently,
\begin{equation}
(1-\epsilon)\|\vct x\| \leq \sqrt{1-\epsilon}\|\vct x\| \leq \|\mat\Psi\vct x\| \leq \sqrt{1+\epsilon}\|\vct x\| \leq (1+\epsilon)\|\vct x\|.
\end{equation}
\end{proof}

\begin{proof}[Proof of Proposition \ref{prop_wellbehave_unisample}]
Let $\Omega\subseteq \{1,\cdots, d\}$, $|\Omega| = p$ be the subsampling indices of $\mat\Omega$.
By definition, $\Pr[\Omega(j) = i] = 1/d$ for every $i\in\{1,\cdots,d\}$ and $j\in\{1,\cdots,p\}$.
Fix any subspace $\mathcal U\subseteq\mathbb R^d$ of dimension at most $r'$ with incoherence level bounded by
$\mu(\mathcal U)\leq\mu_0$.
Let $\mat U\in\mathbb R^{d\times r'}$ be an orthonormal basis of $\mathcal U$.
By definition, $\mat U^\top\mat U=\mat I_{r'\times r'}$.

For any $\vct x\in\mathcal U$, there exists $\vct\alpha\in\mathbb R^{r'}$ such that $\vct x=\mat U\vct\alpha$.
Subsequently, we have
\begin{eqnarray}
\big|\|\vct x\|^2 - \|\mat\Omega\vct x\|^2\big|
= \big|\vct\alpha^\top\vct\alpha - \vct\alpha^\top(\mat\Omega\mat U)^\top(\mat\Omega\mat U)\vct\alpha\big|
\leq \|\vct\alpha\|^2\cdot\|\mat I-(\mat\Omega\mat U)^\top(\mat\Omega\mat U)\|.
 \label{eq_sampling_difference}
\end{eqnarray}

Our next objective is to bound the norm $\|\mat I-(\mat\Omega\mat U)^\top(\mat\Omega\mat U)\|$ with high probability.
First let $\mat U_{\Omega} := (\vct u_{\Omega(1)},\cdots,\vct u_{\Omega(p)}) = \sqrt{\frac{p}{d}}(\mat\Omega\mat U)^\top$ 
be the unnormalized version of subsampled orthogonal operators.
By definition we have
\begin{equation}
\|(\mat\Omega\mat U)^\top(\mat\Omega\mat U) - \mat I\| = \frac{d}{p}\left\|\mat U_{\Omega}\mat U_{\Omega}^\top - \frac{p}{d}\mat I\right\|.
\label{eq_sampling_unnormalized}
\end{equation}

With Eq. (\ref{eq_sampling_unnormalized}), we can use non-commutative Matrix Bernstein inequality \citep{nc-bernstein-1,nc-bernstein-2}
 to bound $\|\mat U_{\Omega}\mat U_{\Omega}^\top - \frac{p}{d}\mat I\|$
and subsequently obtain an upper bound for the rightmost term in Eq. (\ref{eq_sampling_difference}).
The proof is very similar to the one presented in \citep{subspace-detection,adaptive-matrix-completion-arxiv},
where an upper bound for $\|(\mat U_{\Omega}\mat U_{\Omega}^\top)^{-1}\|$ is obtained.
More specifically, let $\mat B_1,\cdots,\mat B_p$ be i.i.d. random matrices such that $\mat B_j = \vct u_{\Omega(j)}\vct u_{\Omega(j)}^\top - \frac{1}{d}\mat I$.
We then have
\begin{equation}
\mat U_{\Omega}\mat U_{\Omega}^\top - \frac{p}{d}\mat I = \sum_{j=1}^p{\mat B_j}
\end{equation}
and furthermore,
\begin{equation}
\mathbb E\left[\mat U_{\Omega}\mat U_{\Omega}^\top - \frac{p}{d}\mat I\right] = p\left(\sum_{i=1}^d{\vct u_i\vct u_i^\top} - \mat I\right) = \mat 0.
\end{equation}
To use Matrix Bernstein, we need to upper bound the range and variance parameters of $\mat B_j$.
Under the matrix incoherence assumption Eq. (\ref{eq_mat_incoherence}) the range of $\mat B_j$ can be bounded as
\begin{equation}
\|\mat B_j\| \leq \max_i{\left\|\vct u_i\vct u_i^\top - \frac{1}{d}\mat I\right\|} \leq \frac{\sqrt{r'^2}\mu_0}{d} + \frac{1}{d} \leq \frac{r'\mu_0}{d} + \frac{1}{d} \leq \frac{2r'\mu_0}{d} =: R.
\end{equation}
The last inequality is due to the fact that $1\leq \mu(\mat U)\leq\frac{d}{r'}$ for any subspace $\mathcal U$ of rank $r'$.
For the variance, we have
\begin{eqnarray*}
\|\mathbb E[\mat B_j^\top\mat B_j]\| = \|\mathbb E[\mat B_j\mat B_j^\top]\|
&=& \left\|\mathbb E\left[\left(\vct u_{\Omega(j)}\vct u_{\Omega(j)}^\top - \frac{1}{d}\mat I\right)\left(\vct u_{\Omega(j)}\vct u_{\Omega(j)}^\top-\frac{1}{d}\mat I\right)\right]\right\|\\
&=& \left\|\mathbb E\left[\vct u_{\Omega(j)}\vct u_{\Omega(j)}^\top\vct u_{\Omega(j)}\vct u_{\Omega(j)}^\top\right] - \frac{1}{d^2}\mat I\right\|\\
&\leq& \left\|\mathbb E\left[\vct u_{\Omega(j)}\vct u_{\Omega(j)}^\top\vct u_{\Omega(j)}\vct u_{\Omega(j)}^\top\right]\right\| + \frac{1}{d^2}\\
&\leq& \frac{\mu_0 \sqrt{r'^2}}{d^2}\|\mathbb E[\vct u_{\Omega(j)}\vct u_{\Omega(j)}^\top]\| + \frac{1}{d^2}\\
&\leq& \frac{\mu_0 r'}{d^2} + \frac{1}{d^2} \leq \frac{2\mu_0 r'}{d^2}.
\end{eqnarray*}
As a result, we can define $\sigma^2 := 2\mu_0 r'/d^2$ such that $\sigma^2 \geq \max\{\|\mathbb E[\mat B_j\mat B_j^\top]\|, \|\mathbb E[\mat B_j^\top\mat B_j]\|\}$ for every $j$.
Using Lemma \ref{lem_nc_bernstein}, for every $t>0$ we have
\begin{equation}
\Pr\left[\|\mat U_{\Omega}\mat U_{\Omega}^\top - \frac{p}{d}\mat I\| \geq t\right] \leq 2r\exp\left(-\frac{t^2/2}{\sigma^2p + Rp/3}\right)
= 2r'\exp\left(-\frac{t^2/2}{\frac{2\mu_0 r'}{d^2}p + \frac{2\mu_0 r'}{d}t/3}\right).
\end{equation}
For $\epsilon < 1$ set $t = \frac{p}{d}\epsilon$ and $p=8\epsilon^{-2}\mu_0 r'\log(2r'/\delta)$.
Then with probability $\geq 1-\delta$ we have
\begin{equation}
\left\|\mat U_{\Omega}\mat U_{\Omega}^\top - \frac{p}{d}\mat I\right\| \leq \frac{p}{d}\epsilon.
\label{eq_sampling_final}
\end{equation}
The proof is then completed by multiplying both sides in Eq. (\ref{eq_sampling_final}) by $\frac{d}{p}$.

\end{proof}

\section{Some tail inequalities}\label{app:tailbounds}

\begin{lem}[Matrix Gaussian and Rademacher Series, the general case \cite{mat-friendly}]
Let $\{\mat B_k\}_k$ be a finite sequence of fixed matrices with dimensions $d_1\times d_2$.
Let $\{\gamma_k\}_k$ be a finite sequence of i.i.d. standard normal variables.
Define the summation random matrix $\mat Z$ as
\begin{equation}
\mat Z = \sum_k{\gamma_k\mat B_k}.
\end{equation}
Define the variance parameter $\sigma^2$ as
\begin{equation}
\sigma^2 := \max\{\|\mathbb E[\mat Z\mat Z^\top]\|, \|\mathbb E[\mat Z^\top\mat Z]\|\}.
\end{equation}
Then for every $t>0$ the following concentration inequality holds:
\begin{equation}
\Pr\left[\|\mat Z\| \geq t\right] \leq (d_1+d_2)e^{-t^2/2\sigma^2}.
\end{equation}
\label{lem_gaussian_sigma1}
\end{lem}

\begin{lem}[Noncommutative Matrix Berstein Inequality, \cite{nc-bernstein-1,nc-bernstein-2}]
Let $\mat B_1,\cdots,\mat B_p$ be independent zero-mean square $r\times r$ random matrices.
Suppose $\sigma_j^2 = \max\{\|\mathbb E[\mat B_j\mat B_j^\top]\|, \|\mathbb E[\mat B_j^\top\mat B_j]\|\}$
and $\|\mat B_j\|\leq R$ almost surely for every $j$.
Then for any $t>0$ the following inequality holds:
\begin{equation}
\Pr\left[\left\|\sum_{j=1}^p{\mat B_j}\right\|_2 > t\right] \leq 2r\exp\left(-\frac{t^2/2}{\sum_{j=1}^p{\rho_j^2} + Rt/3}\right).
\end{equation}
\label{lem_nc_bernstein}
\end{lem}

\begin{lem}[Spectrum bound of a Gaussian random matrix,\cite{davidson2001local}]\label{lem_gaussian_spectrum}
Let $A$ be an $m\times n$ $(m> n)$ matrix with i.i.d standard Gaussian entries. Then, its largest and smallest singular values $s_1(A)$ and $s_n(A)$ obeys
$$
\sqrt{m}-\sqrt{n} \leq \E s_n(A) \leq \E s_1(A) \leq \sqrt{m}+\sqrt{n},
$$
moreover,
$$
\sqrt{m}-\sqrt{n}-t \leq s_n(A) \leq s_1(A) \leq \sqrt{m}+\sqrt{n}+t,
$$
with probability at least $1-2\exp(-t^2/2)$ for all $t>0$.
\end{lem}
The expectation result is due to Gordon's inequality and the concentration follows from the concentration of measure inequality in Gauss space by the fact that $s_1$ and $s_n$ are both 1-Lipchitz functions. Take $t=\sqrt{n}$ in the above inequality we get
$$
1-2\sqrt{\frac{n}{m}} - \epsilon \leq s_n(A/\sqrt{m}) \leq s_1(A/\sqrt{m}) \leq 1 +2\sqrt{\frac{n}{m}}
$$
with probability $1-2\exp(-n^2/2)$.

\end{appendices}

\bibliographystyle{apa-good}
\bibliography{sc-random}

\clearpage

\section*{Table of symbols and notations}

\begin{table}[!htb]
\centering
\caption{Summary of symbols and notations}
\scalebox{0.8}
{
\begin{tabular}{ll}
\hline
$|\cdot|$& Either absolute value or cardinality\\
$\|\cdot\|$; $\|\cdot\|_2$& 2 norm of a vector/spectral norm of a matrix\\
$\|\cdot\|_1$& 1 norm of a vector\\
$\|\cdot\|_{\infty}$& Infinity norm (maximum absolute value) of a vector\\
$\langle\cdot,\cdot\rangle$& Inner product of two vectors\\
$\|\mat A\|_{(i)}$& The $i$th row of matrix $\mat A$\\
$\sigma_1(\cdot),\sigma_r(\cdot)$& The largest and $r$th largest singular value of a matrix\\
$N$& Number of data points (number of columns in $\mat X$)\\
$k$& Number of subspaces (clusters)\\
$d$& The ambient dimension (number of rows in $\mat X$)\\
$N_\ell,r_\ell$ for $\ell=1,\cdots,k$& Number of data points and intrinsic dimension for each subspace\\
$r$& Largest intrinsic dimension across all subspaces\\
$\mat X$& Observed data matrix\\
$\mat Y$& Uncorrupted (noiseless) data matrix\\
$\mat Z$& Noise matrix, can be either deterministic or stochastic\\
$\widetilde{\mat X},\widetilde{\mat Y},\widetilde{\mat Z}$& Projected matrices of $\mat X,\mat Y,\mat Z$\\
$\widetilde{\mat X}',\widetilde{\mat Y}',\widetilde{\mat Z}'$& Normalized projected matrices of $\mat X,\mat Y,\mat Z$\\
$\mathcal U^{(\ell)}$, $\mat U^{(\ell)}$& Subspace and its orthonormal basis of the $\ell$th cluster\\
$\mat X_{-i}$, $\mat Y_{-i}$, $\mat Z_{-i}$& All columns in $\mat X,\mat Y,\mat Z$ except the $i$th column.\\
$\mat X^{(\ell)}$,$\mat Y^{(\ell)}$,$\mat Z^{(\ell)}$& All columns in $\mat X,\mat Y,\mat Z$ associated with the $\ell$th subspace\\
$\mat X_{-i}^{(\ell)}$,$\mat Y_{-i}^{(\ell)}$,$\mat Z_{-i}^{(\ell)}$& All columns in $\mat X^{(\ell)},\mat Y^{(\ell)},\mat Z^{(\ell)}$ except the $i$th column\\
$\mathcal Q(\cdot)$,$\text{conv}(\cdot)$& (Symmetric) convex hull of a set of vectors\\
$r(\cdot)$& Radius of the largest ball inscribed in a convex body\\
$\mathcal P_{\mathcal U}(\cdot)$& Projection onto subspace $\mathcal U$\\
$p$& Target dimension after random projection\\
$\epsilon$& Approximation error of random projection methods\\
$\delta$& Failure probability\\
$\mat\Psi,\mat\Omega,\mat\Phi,\mat S$& Projection operators for random Gaussian projection, uniform sampling, FJLT and sketching\\
$\mu_0$& Column space incoherence or column spikiness\\
$\mu_\ell,\rho_\ell$ for $\ell=1,\cdots,k$& Subspace incoherence and inradius for each subspace\\
$\tilde\mu_\ell,\tilde\rho_\ell$ for $\ell=1,\cdots,k$& Subspace incoherence and inradius on the projected data\\
$f(\cdot),\tilde f(\cdot)$& Objective functions  of Eq. (\ref{eq_lasso_dual}) on the original data and projected data\\
$\vct\nu,\vct v$& Unnormalized and normalized dual direction\\
$\tilde{\vct\nu}$& Random projection of $\vct\nu$\\
$\tilde{\vct\nu}'$& A shrunken version of $\tilde{\vct\nu}$ such that it is feasible for Eq. (\ref{eq_lasso_dual}) on projected data\\
$\vct\nu^*$& Optimal solution to Eq. (\ref{eq_lasso_dual}) on projected data\\
$\bar{\vct\nu}$& A vector in the original space that corresponds to $\vct\nu^*$ after projection\\
$\bar{\vct\nu}'$& A shrunken version of $\bar{\vct\nu}$ such that it is feasible for Eq. (\ref{eq_lasso_dual}) on the original data\\
$\lambda$& Regularization coefficient for Lasso SSC\\
$\Delta$& Margin of error (i.e., $\min_{\ell}{\rho_\ell-\mu_\ell}$)\\
$\eta, \tilde\eta$& Noise level for deterministic noise, before and after projection\\
$\sigma,\tilde\sigma$& Noise level for random Gaussian noise, before and after projection\\
$\mat C$& Similarity matrix\\
$q$& Number of nonzero entries in regression solutions. Used in solution path algorithms.\\
\hline
\end{tabular}
}
\end{table}

\end{document}